\documentclass[lettersize,journal]{IEEEtran}
\usepackage{amsmath,amsfonts}
\usepackage{algorithmic}
\usepackage{algorithm}
\usepackage{array}
\usepackage[caption=false,font=normalsize,labelfont=sf,textfont=sf]{subfig}
\usepackage{textcomp}
\usepackage{stfloats}
\usepackage{url}
\usepackage{verbatim}
\usepackage{graphicx}
\usepackage{cite}
\usepackage{multirow}
\usepackage{array}
\usepackage[T1]{fontenc}
\usepackage{booktabs}
\usepackage{amssymb}
\usepackage{pifont}
\usepackage[numbers,sort&compress]{natbib}
\usepackage{makecell}
\makeatletter

\newcommand{\Rmnum}[1]{\expandafter\@slowromancap\romannumeral #1@}
\makeatother
\usepackage{threeparttable}
\usepackage{xcolor} 
\usepackage{bigstrut}

\begin{document}

\title{RingMo-lite: A Remote Sensing Multi-task Lightweight Network with CNN-Transformer Hybrid Framework}

\author{ Yuelei~Wang, 
Ting~Zhang,
Liangjin~Zhao,
Lin~Hu,
Zhechao~Wang,
Ziqing~Niu,
Peirui~Cheng,
Kaiqiang~Chen,
Xuan~Zeng, 
Zhirui~Wang, 
Hongqi~Wang, \IEEEmembership{Member,~IEEE},
and Xian~Sun, \IEEEmembership{Senior Member,~IEEE} 
\thanks{This work was supported by the National Key R\&D Program of China (2022ZD0118402), and the National Nature Science Foundation of China under Grant 62331027 and Grant 62076241.
\textit{(Corresponding author: Xian Sun.)}}
\thanks{Yuelei Wang, Ting Zhang, Lin Hu, Zhechao Wang, Ziqing Niu, Xuan Zeng and Xian Sun are with the Aerospace Information Research Institute, Chinese Academy of Sciences, Beijing 100190, China, also with the Key Laboratory of Network Information System Technology (NIST), Aerospace Information Research Institute, Chinese Academy of Sciences, Beijing 100190, China, also with the University of Chinese Academy of Sciences, Beijing 100190, China, and also with the School of Electronic, Electrical and Communication Engineering, University of Chinese Academy of Sciences, Beijing 100190, China (e- mail: wangyuelei20@mails.ucas.ac.cn; zhangting20@mails.ucas.ac.cn;  hulin21@mails.ucas.ac.cn; wangzhechao21@mails.ucas.ac.cn;  niuziqing21@mails.ucas.ac.cn; zengxuan19@mails.ucas.ac.cn; sunxian@aircas.ac.cn). }
\thanks{Liangjin Zhao, Peirui Cheng, Kaiqiang Chen, Zhirui Wang and Hongqi Wang are with the Aerospace Information Research Institute, Chinese Academy of Sciences, Beijing 100094, China, and also with the Key Laboratory of Network Information System Technology (NIST), Aerospace Information Research Institute, Chinese Academy of Sciences, Beijing 100190, China (e-mail: zhaolj004896@aircas.ac.cn; chengpr@aircas.ac.cn; chenkq@aircas.ac.cn; zhirui1990@126.com; wiecas@sina.com).}
}

\maketitle

\begin{abstract}
In recent years, remote sensing (RS) vision foundation models such as RingMo have emerged and achieved excellent performance in various downstream tasks. However, the high demand for computing resources limits the application of these models on edge devices. It is necessary to design a more lightweight foundation model to support on-orbit RS image interpretation. Existing methods face challenges in achieving lightweight solutions while retaining generalization in RS image interpretation. This is due to the complex high and low-frequency spectral components in RS images, which make traditional single CNN or Vision Transformer methods unsuitable for the task. Therefore, this paper proposes RingMo-lite, an RS multi-task lightweight network with a CNN-Transformer hybrid framework, which effectively exploits the frequency-domain properties of RS to optimize the interpretation process. It is combined by the Transformer module as a low-pass filter to extract global features of RS images through a dual-branch structure, and the CNN module as a stacked high-pass filter to extract fine-grained details effectively. Furthermore, in the pretraining stage, the designed frequency-domain masked image modeling (FD-MIM) combines each image patch's high-frequency and low-frequency characteristics, effectively capturing the latent feature representation in RS data. As shown in Fig. \ref{fig_compare}, compared with RingMo, the proposed RingMo-lite reduces the parameters over 60\% in various RS image interpretation tasks, the average accuracy drops by less than 2\% in most of the scenes and achieves SOTA performance compared to models of the similar size. In addition, our work will be integrated into the MindSpore computing platform in the near future.

\end{abstract}

\section{Introduction}

\begin{figure}
\centering
	\includegraphics[scale=.04]{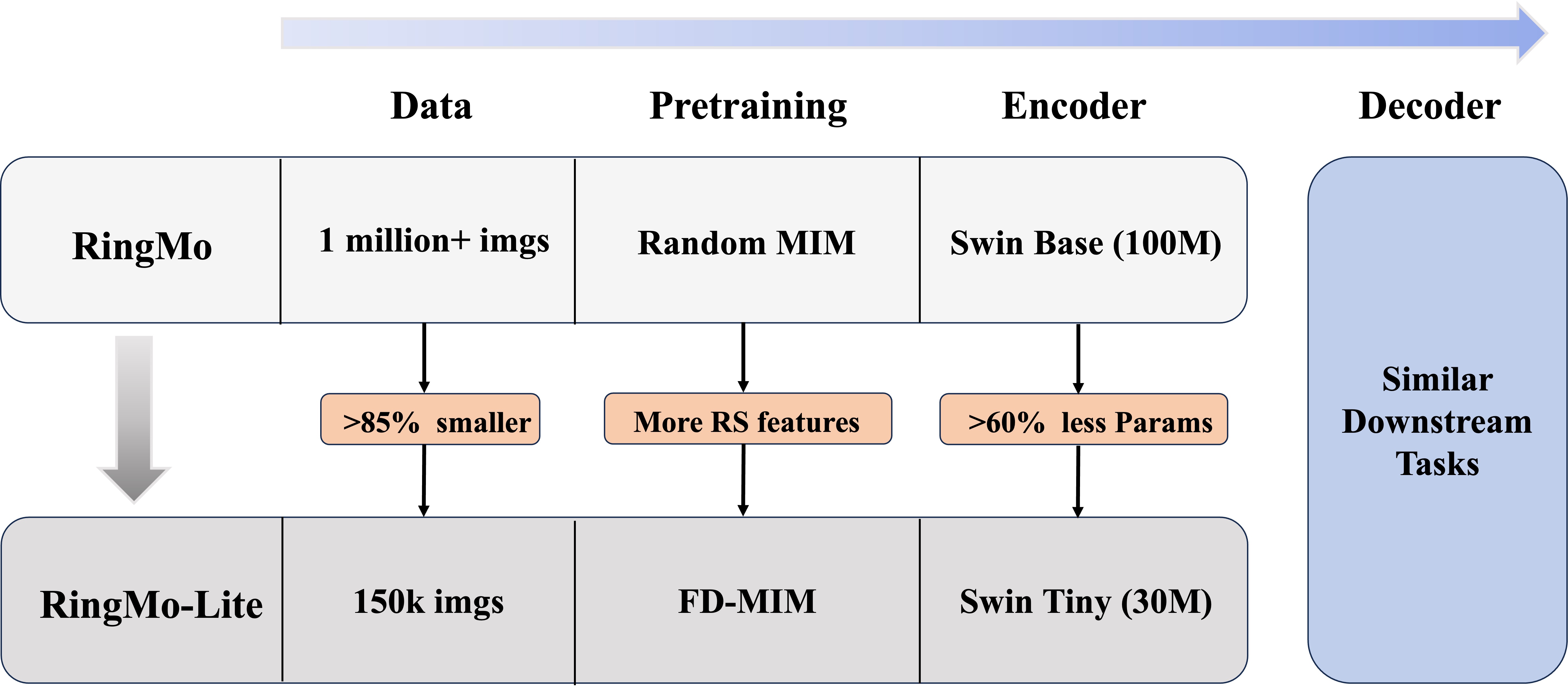}
\caption{Comparison of RimgMo and our proposed RingMo-lite  from four dimensions: data (use 85\% smaller image datasets for pretraining MIM), pretraining methods (use frequency domain for more RS features), encoder (over 60\% fewer parameters) and decoder (similar computer vision tasks).}
\label{fig_compare}
\end{figure}

\begin{figure*}
\centering
	\includegraphics[scale=.45]{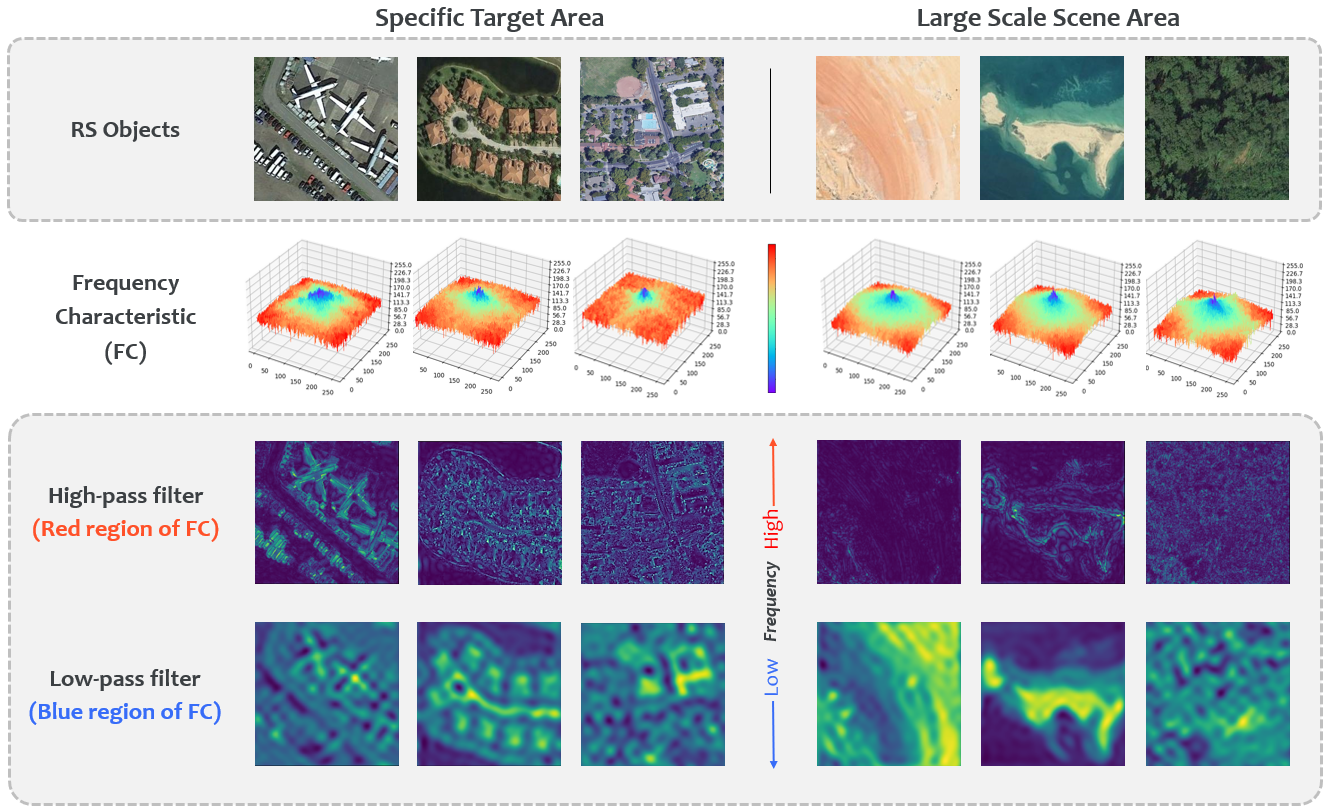}
\caption{Examples of frequency-domain comparison between specific target area and large-scale scene area in different RS scenes. The 3D frequency domain diagram in the second row is calculated based on the spectral components, where the closer to the center represents the low-frequency part, and the closer to the periphery represents the high-frequency part. The third and fourth rows are the results of the image after high-pass filter and low-pass filter respectively.
}
\label{fig_1}
\end{figure*}

\IEEEPARstart{I}{n} recent years, many Transformer-based methods \cite{liu2021swin, dong2023distilling, chen2021remote} have emerged and achieved great success with their excellent feature extraction and representation capabilities, especially for vision foundation models \cite{dosovitskiy2020image}. Unlike the unsupervised method \cite{deng2021joint}, the foundation model shows an extreme capacity for generalization. In the field of remote sensing (RS), RingMo \cite{sun2022ringmo}, the RS foundation model has been proposed, which effectively solves the problem of inadequate generalization ability for existing methods. However, the foundation models are not flexible and efficient due to the high demand for computing and storage resources \cite{mehta2021mobilevit,liu2021vision,chen2022mobile}, making it difficult to adapt to edge servers or terminals, and cannot support on-orbit RS image interpretation in practical applications. Therefore, designing a lightweight foundation model and achieving high-precision, multi-task on-orbit interpretation of RS images is a future development trend.

Many methods have been proposed to realize lightweight vision foundation models in the field of general visual processing \cite{cai2022efficientvit,pan2022edgevits,ma2022mocovit}. These methods can be categorized into three groups: knowledge distillation \cite{wu2022tinyvit, liu2022cross}, neural architecture search (NAS) \cite{chen2021glit, wang2020hat,xu2021bert}, and network structure design \cite{wadekar2022mobilevitv3,zhang2022edgeformer}. Among them, knowledge distillation transfers knowledge from the large-scale foundation model to the student model, compressing the size of the model while maintaining its performance. However, this approach needs training an additional teacher model and requires separate distillations for various downstream tasks. NAS is another approach to achieve model lightweight, which automatically explores different neural network structures to find a model with good performance and fewer parameters. However, it demands significant computational resources and processing time. The network structure design methods achieve a significant parameter reduction by changing the transformer block structure of foundation models, without training additional models and saving computing resources. 

Although network structure design offers a viable option, there are still two challenges in the field of RS that may affect the performance of the method. Firstly, as shown in Fig. \ref{fig_1}, RS images have various resolution and orientation ranges, and the distribution of objects is complex \cite{cheng2020remote,sun2020bas,wang2023sar,xian2019air}. Therefore, RS images usually contain specific target areas and large-scale ground objects simultaneously, with many scale differences between them. The pixels of dense small objects change drastically in the spatial dimension, while the pixels of large-scale ground objects change relatively more uniformly and slowly. The multi-scale differences in these objects bring great challenges to the generalization ability of the model. Secondly, various RS interpretation tasks prefer to focus on different target areas. For example, the scene classification task \cite{lu2021lil,zhang2021cross,zhao2020joint} involves a wide range of spatial scales, thus requiring more attention to the global generalization information. However, in the downstream tasks of RS object detection \cite{sun2021pbnet,zhu2020diamondnet}, it is necessary to pay more attention to the local detail information of targets such as aircraft, ships and vehicles. The pixel changes of critical objects in RS images have corresponding representations in the frequency domain, and different frequencies refer to the intensity of feature changes. These differences in the high-frequency and low-frequency information affect the interpretation accuracy of different downstream tasks to a certain extent. Although many network structure design methods adopt a combination of CNN \cite{he2016deep,sandler2018mobilenetv2} and Transformer \cite{vaswani2017attention}, they mainly focus on using CNN to replace parts of the Transformer block to reduce calculations. Most existing methods do not pay attention to the advantages of using CNN and Transformer in extracting high-frequency and low-frequency information from RS images.

Considering the above problems, this paper proposes a novel lightweight foundation model RingMo-lite suitable for various RS image interpretation tasks. Firstly, in order to fully extract the detailed features of specific target areas and the global features of large-scale scenes, this paper designs a lightweight CNN-Transformer dual-branch hybrid architecture. Specifically, the Transformer structure establishes the global relationship and long-distance dependence through the self-attention mechanism, which enables a more profound comprehension of both the structural and semantic aspects of the image. Therefore, in the frequency domain of the input image, Transformer can be regarded as the low-pass filter to extract low-frequency information, which can better extract the information of large-scale surface feature elements. On the contrary, the CNN architecture pays attention to the local details in the convolution sliding window through matrix calculation. Thus, the CNN branch aims to further alleviate the spatial position bias and capture local features such as texture and details. In the frequency domain, CNN can be regarded as the superposition of multiple high-pass filters, which is more suitable for extracting high-frequency information and processing specific target information. Combining the advantages of the two different structures of CNN and Transformer, the proposed dual-branch block decouples the hybrid structure in the channel dimension, which comprehensively utilizes high-frequency and low-frequency information in RS images and effectively improves the interpretation accuracy.
\IEEEpubidadjcol

Secondly, this paper designs a frequency domain masked image modeling (FD-MIM) that adapts to high-frequency and low-frequency information of RS images, which improves the pretraining effect of lightweight foundation models by combining self-supervised learning  \cite{he2022masked,xie2022simmim}. The FD-MIM, which corresponds to the proposed CNN-Transformer hybrid framework, contributes to better reconstruction of image details during masking, and promotes the proposed lightweight model to learn a rich feature representation suitable for different downstream tasks.

This paper’s contribution can be summarized as follows:

\begin{enumerate}
\item{In order to achieve lightweight on-orbit interpretation, this paper proposes RingMo-lite, a dual-branch CNN-Transformer hybrid framework suitable for various RS image interpretation tasks. The proposed method fully considers the high-frequency and low-frequency information of RS images and tasks and effectively improves interpretation accuracy.}

\item{Considering the frequency-domain characteristics of RS object areas, this paper designs an FD-MIM self-supervised pretraining strategy, which facilitates the proposed framework to learn richer feature representations and effectively improves the generalization ability in downstream tasks.}

\item{Compared with RingMo, RingMo-lite reduces the parameters over 60\% in various RS image interpretation tasks, the average accuracy drops by less than 2$\%$, and  RingMo-lite achieves SOTA performance in four downstream tasks compared to models of the same scale, including RS image classification, object detection, semantic segmentation, and change detection.}

\end{enumerate}

In addition, we are planning to integrate Ringmo-lite into Huawei AI chips, which has the MindSpore computing platform, to enable deployment on edge devices.

The rest part of the paper is recognized as follows. Section \ref{relate} will introduce the related work. Section \ref{method} will present the RingMo-lite algorithm, and Section \ref{exp} will discuss several experiments of the RingMo-lite performance. In Section \ref{Con}, the conclusion will be given.

\section{Related Work}\label{relate}

\subsection{Lightweight Foundation Models in General Vision Domain}
As the two dominant architectural paradigms in the field of computer vision, CNNs and ViTs have garnered significant attention and research in recent years due to their remarkable performance in tasks such as image classification, object detection, semantic segmentation and change detection. CNNs excel at capturing local features and high-frequency information owing to their advantages of local receptive fields, parameter sharing, and hierarchical feature extraction, thus establishing them as the classical method for image processing. In contrast, ViTs leverage the self-attention mechanism of Transformer \cite{vaswani2017attention}, which is good at capturing global relationships, and exhibit immense potential in the image domain. However, the exceptional performance of both architectures critically depends on their substantial parameters and computational consumption, limiting their deployment on resource-constrained devices. To address this issue, a multitude of research endeavors focusing on model lightweight have emerged, aiming to maintain high performance while reducing computational resources and memory requirements. In the realm of lightweight CNNs, pioneering achievements like MobileNet \cite{howard2017mobilenets}, EfficientNet \cite{tan2019efficientnet}, and ShuffleNet \cite{zhang2018shufflenet}, have harnessed innovations such as depth-wise separable convolutions and channel attention. In the domain of lightweight ViTs, methodologies like MobileViT \cite{mehta2021mobilevit}, TinyViT \cite{wu2022tinyvit}, and EfficientViT \cite{cai2023efficientvit} have emerged, which mainly concentrate on techniques such as knowledge distillation and model pruning to compress ViT models. This paper will introduce existing lightweight strategies around knowledge distillation, quantization and pruning, and Neural Architecture Search (NAS). These methods collectively propel the efficient deployment of deep learning models in edge devices.

\textbf{Knowledge Distillation-Based Lightweight Method:} This approach involves transferring rich knowledge (including logits, intermediate features, etc.) from large pretrained models to small student models. Wu \emph{et al.} \cite{wu2022tinyvit} introduce TinyViT, achieving fast distillation by sparsifying logits of the large teacher model in advance and storing them, which saves forward computations and memory usage. Liu \emph{et al.} \cite{liu2022cross} propose a cross-architecture knowledge distillation method to distill complementary knowledge from Transformer to guide the training of CNNs. Yang \emph{et al.} \cite{yang2022vitkd} introduce ViTKD, which utilizes the nature of feature maps in ViT to design a knowledge distillation method suitable for the ViT structure. Liang \emph{et al.} \cite{liang2023less} propose TED, a method that aligns the hidden representations of the student and teacher model at each layer by designing task-aware filters, selecting knowledge beneficial for the target task, and reducing the knowledge gap between the two models to help the student model better adapt to the target task.

\textbf{Quantization and Pruning-Based Lightweight Method:} Yu \emph{et al.} \cite{yu2022unified} propose an end-to-end lightweight framework that combines pruning, skip connections, and distillation, significantly reducing computational complexity while nearly maintaining model performance. Yin \emph{et al.} \cite{yin2023gohsp} introduce GOHSP, which leverages graph-based ranking methods to measure the importance of attention heads and then integrates the extracted importance information into an optimization-based scheme to induce heterogeneous structure sparsity in ViT models. Wei \emph{et al.} \cite{wei2023joint} propose TPS, using one-way nearest neighbor matching and similarity-based fusion to combine pruned token information with retained tokens, which mitigates performance degradation caused by pruning.

\textbf{NAS-based lightweight method:} In the current mainstream lightweight method, NAS generates a mobile terminal model by searching and stacking small basic units, and automatically implements network model design, which has obvious advantages. Google proposes MnasNet \cite{tan2019mnasnet}, which considers the real model delay parameters in the neural network structure search process, proposes a decomposed hierarchical search space, and obtains a deep neural model that achieves an optimal balance between model accuracy and model delay. In the exploration of lightweight large models, due to serious conflicts between the gradients of different subnets and supernets, the training is prone to early saturation and poor convergence. Therefore, the direct application of supernet-based NAS for optimization will lead to performance degradation. To this end, NASVIT \cite{gong2021nasvit} proposes a series of techniques, including a gradient projection algorithm, a switchable layer scaling design, and a simplified data augmentation and regularization training recipe, which effectively improve the convergence and performance of the subnetwork.

The aforementioned lightweight methods have been extensively validated in general vision domain. However, unlike natural images, RS images exhibit characteristics such as complex object distributions and diverse spatial scales, which leads to the limited applicability of the above general methods in RS tasks. Therefore, it is necessary to design a more suitable method for the characteristics of RS images.

\subsection{Lightweight Networks in Remote Sensing Domain}
Different from natural scene images, due to the essential differences in the scale and direction of the objects generated by the bird's-eye view, RS images have complex backgrounds and dense targets, which makes intelligent interpretation more difficult.

RS scene classification is widely used in the RS community, but traditional CNN-based methods lack long-range dependencies and cannot fully capture contextual information, while fine-tuning pretrained large models, such as ViT, is costly. To this end, LTNet \cite{huang2023faster} proposes a lightweight transformer network, which captures global dependencies with low computing resources through the multi-level group convolution (MLGC) module, improves the diversity of local features, and enhances classification performance. In order to solve the problem of limited links between ViT adjacent windows and huge computational load, LDBST \cite{zheng2023lightweight} designs a dual-branch structure combining ViT branches and CNN branches based on the hierarchical Swin Transformer model. The ViT branch promotes the connection of adjacent windows through the dual multi-layer perceptron structure with deep convolutional layers, and the CNN branch with maximum pooling not only retains the discriminative ability of scene features but also avoids the huge computation caused by complex multi-head attention. However, it does not consider distinguishing high-frequency information from low-frequency information, which affects the interpretation accuracy to a certain extent.

As a more difficult task, object detection suffers from stronger interference in RS scenarios. Aimed at the problem of complex RS image background and difficult detection of small targets in dense scenes, Chen \emph{et al.} \cite{chen2023mdct} proposes a single-stage object detection model called MDCT based on a multi-kernel dilated convolution block (MDC) and Transformer block, in which the MDC module is used to enhance the features of small targets and increase the receptive field. The transformer block is integrated into the neck network of the detection model to prevent the loss of object information in complex backgrounds. Considering that the introduction of the attention mechanism will lead to an increase in computational complexity as the image resolution increases, Gong \emph{et al.} \cite{gong2022swin} proposes to replace the convolutional prediction head with Swin Transformer prediction heads (SPH). The designing of the shift window effectively reduces the computational complexity, while introducing a feature fusion layer to preserve feature information to the maximum extent.

In RS semantic segmentation, it is also limited by the ability of CNN to capture the global context, and the segmentation accuracy reaches a bottleneck. To address this issue, Wang \emph{et al.} \cite{wang2022unetformer} proposes a hybrid architecture UNetFormer consisting of a CNN-based encoder and a transformer-based decoder to handle RS semantic segmentation tasks, where the decoder consists of a global-local Transformer block (GLTB) to efficiently model global and local information. However, transformer-based architectures usually suffer from high computational load and low-precision edge classification when applied to RS semantic segmentation tasks. To this end, Xu \emph{et al.} \cite{xu2021efficient} proposes a purely efficient lightweight model with MLP head based on Swin Transformer to speed up inference, and deal with the edge problem through explicit and implicit edge enhancement methods.

Although the aforementioned methods of lightweight foundation models in the field of RS have achieved certain results, they have not considered the distinction between high-frequency and low-frequency information in RS images, so as to realize the comprehensive utilization of information. Furthermore, these methods are all trained exclusively on single tasks, lacking abilities for multi-task generalization.

\section{Methodology}\label{method}
\begin{figure*}
\centering
	\includegraphics[scale=.065]{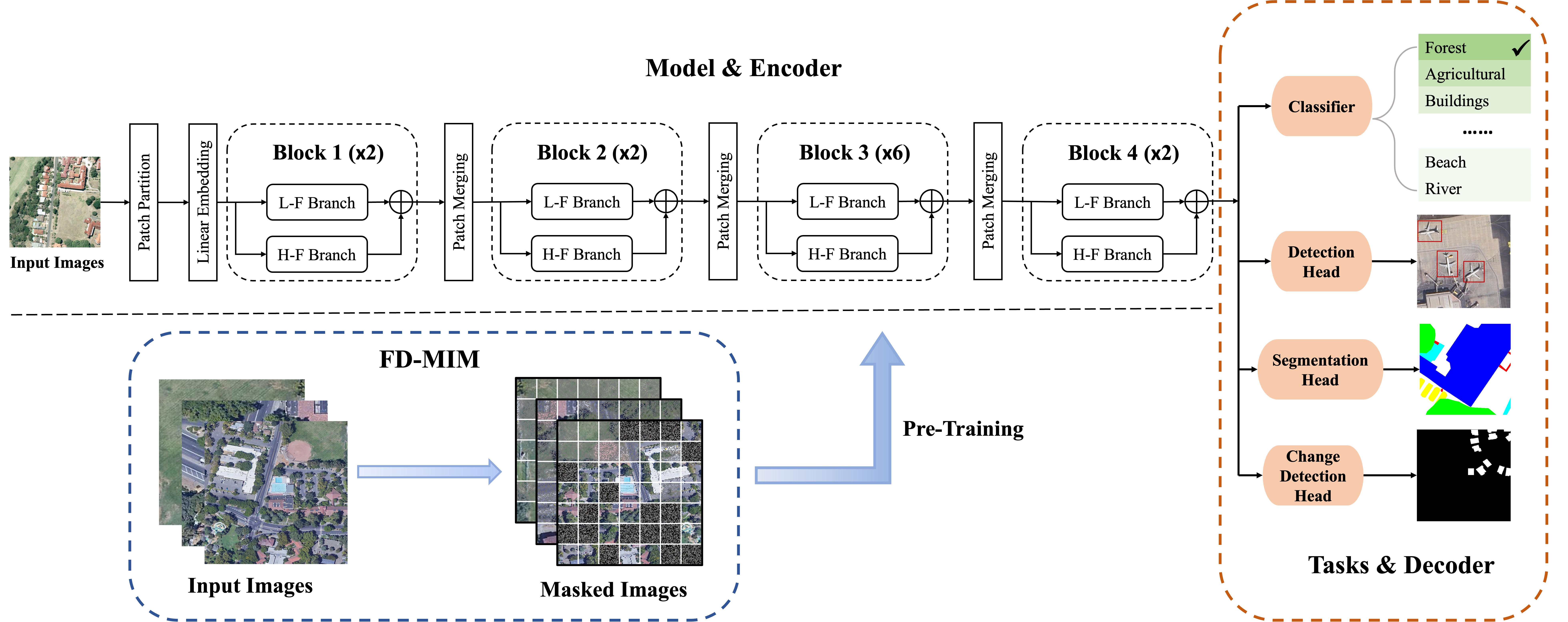}
\caption{The overall architecture of the proposed RingMo-lite. During the pretraining, the FD-MIM strategy is employed to efficiently extract feature representations of RS images. Given an optical RS image as input, the features are first extracted by the encoder composed of several high-low frequency dual-branch blocks (FIFB). Then, four different downstream tasks are achieved through different decoder heads, including classification, object detection, semantic segmentation and change detection from top to bottom.}
\label{fig_overview}
\end{figure*}

\begin{table*}[h!t]
  \centering
  \caption{Model configurations for RingMo-lite. We use the model for classification as an example.}
  \centering
  \label{tab_method}
  \renewcommand\arraystretch{1.3}
  \setlength{\tabcolsep}{6.0mm}{
  \begin{tabular}{c|c|c|c}
  \toprule[1.5pt]

    & Output Size & Layer Name & RingMo-lite \\
   \midrule[0.75pt]
   \multirow{2}{*}[-5.0ex]{Stage 1} & 56 × 56 × 96 & Patch Embedding & $Kernel: 4 \times 4, Stride = 4, Embed\_Dim = 96$ \\
   \cline{2-4}
   \multirow{2}{*}{} & 56 × 56 × 96 &CNN-Transformer Hybrid Block & $\begin{bmatrix}
 num\_heads=3\\
 R=4\\
 w.s.=7\\
FIFB\\
\end{bmatrix} \times 2 $\\
   \midrule[0.75pt]
   Stage 2 & 28 × 28 × 192 &CNN-Transformer Hybrid Block & $\begin{bmatrix}
 num\_heads=6\\
 R=4\\
 w.s.=7\\
FIFB\\
\end{bmatrix} \times 2 $ \\
   
   \midrule[0.75pt]
   Stage 3 & 14 × 14 × 384 &CNN-Transformer Hybrid Block & $\begin{bmatrix}
 num\_heads=12\\
 R=4\\
 w.s.=7\\
FIFB\\
\end{bmatrix} \times 6 $ \\
   
   \midrule[0.75pt]
   Stage 4 & 7 × 7 × 768 &CNN-Transformer Hybrid Block & $\begin{bmatrix}
 num\_heads=24\\
 R=4\\
 w.s.=7\\
FIFB\\
\end{bmatrix} \times 2 $ \\

   \midrule[0.75pt]
   \multicolumn{3}{c|}{Params (M)} & 28.294\\
   \midrule[0.75pt]
   \multicolumn{3}{c|}{FLOPs (G)} & 4.494\\
  
  \bottomrule[1.5pt]
  \end{tabular}}
\end{table*}

\subsection{Overview}
The overall framework of the proposed method is illustrated in Fig \ref{fig_overview}. The input image is initially partitioned into non-overlapping patches using the Patch Partition module, with each patch being the size of 4$\times$4 and treated as a token. These patches are then stacked together as the input to the subsequent linear embedding layer. Subsequently, processed image representations are obtained through four stages. Each stage comprises varying numbers of High-Low Frequency Information Fusion Block (FIFB), with the specific quantity adhering to the configuration of (2, 2, 6, 2) of Swin Tiny \cite{liu2021swin}. As the network deepens, the Patch Merging layer is introduced between stages to counteract the diminishing number of tokens. Within each FIFB, there exists a subdivision into Low-Frequency (L-F) Branch and the High-Frequency (H-F) Branch. To optimally exploit the feature extraction capabilities of both CNNs and Transformers, the input features of FIFB are sent to two branches respectively to capture low-frequency information and high-frequency information, and then fused and fed to the next Block or Patch Merging layer. The L-F Branch follows the main structure of Swin Transformer and obtains global features. H-F Branch further divides the input features into two parts and uses CNNs to extract detailed features.

\subsection{High-Low Frequency Information Fusion Block (FIFB)}

\begin{figure}
\centering
	\includegraphics[scale=.065]{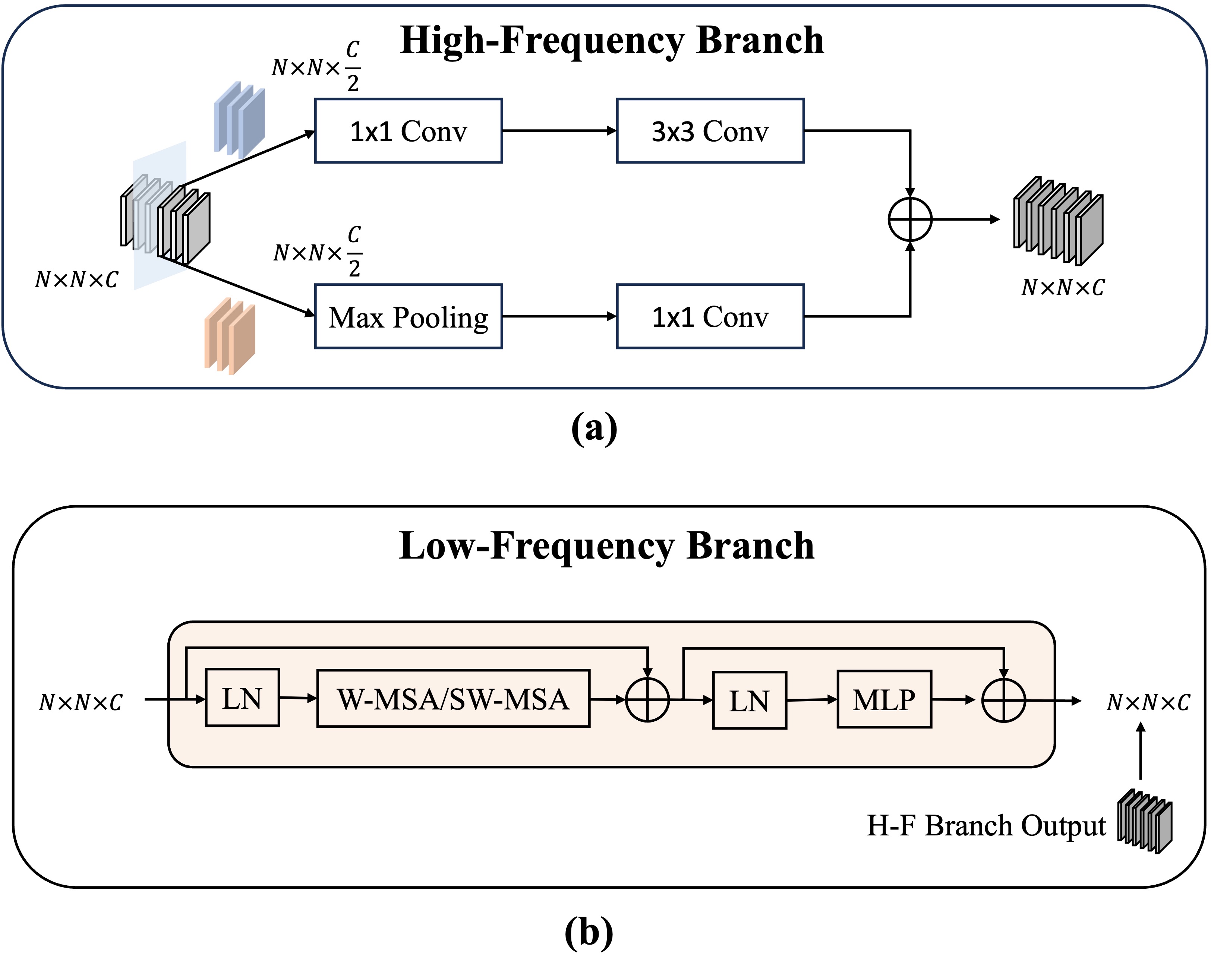}
\caption{The structure of H-F Branch and L-F Branch and illustration of the variation of feature dimension. For H-F Branch, CNNs are used to extract local information. Meanwhile, for L-F Branch, the Swin Transformer blocks are utilized for global information. The features then follows residual computation to the next block.}
\label{fig_h-lf}
\end{figure}

\textbf{Revisiting Vision Transformer and CNNs:} ViT leverages Multi-head Self-Attention (MSA) for information exchange among non-overlapping tokens. As a low-pass filter \cite{park2022vision}, MSA excels in modeling long dependencies and capturing low-frequency information. Nevertheless, MSA's spatial smoothing operations on feature maps tend to attenuate high-frequency signals, leading to feature representations dominated by low-frequency information. Conversely, CNNs employ local convolutions (Convs) within receptive fields to obtain local information. In contrast to MSA, Convs are high-pass filters \cite{park2022vision} and effectively extract high-frequency representations of images. As a result, MSA and Convs exhibit complementary characteristics, with MSA excelling in capturing global dependencies and low-frequency information, while Conv excels in preserving local details and high-frequency information.

\textbf{Frequency Characteristics in Remote Sensing Tasks:} Typically, the global structures of scenes and objects convey low-frequency information in images, while local spatial details such as edges and textures manifest as high-frequency information. RS images inherently encompass both small targets and extensive geographical features. The pixels of densely distributed and small-scale targets change drastically in spatial, while large-scale features are relatively uniform and slow.  As for RS image interpretation tasks, scene classification emphasizes the extraction of comprehensive global information, while object detection tasks concentrate on capturing details. Furthermore, more fine-grained tasks require more local details. In light of these considerations, we propose the FIFB, which combines high-frequency and low-frequency information, thereby promoting the model's ability for multi-task generalization of RS images.

\textbf{FIFB:} As shown in Fig. \ref{fig_h-lf}, the input feature $F \in {\mathbb{R}^{N \times N \times C}}$ of FIFB are separately fed into two distinct branches: the L-F Branch and the H-F Branch. The L-F Branch is based on the architecture of Swin Transformer to capture extensive dependencies over long distances. The input feature $F$ first undergo Layer Normalization (LN) layer and subsequently engage in an alternating sequence of Windowed Multi-Head Self-Attention (W-MSA) and Shifted Windowed Multi-Head Self-Attention (SW-MSA) modules, after which a residual link is applied. An additional LN layer and two Multi-Layer Perceptron (MLP) layers are further connected. After that, another residual connection is employed to obtain the output of the low-frequency branch, which is defined as $L$.
\begin{equation}
\label{eq_1}
\hat L = {{\mathop{\rm MSA}\nolimits} _{W/SW}}\left( {{\mathop{\rm LN}\nolimits} \left( F \right)} \right) + F
\end{equation}

\begin{equation}
\label{eq_2}
L = {\mathop{\rm MLP}\nolimits} \left( {{\mathop{\rm LN}\nolimits} \left( {\hat L} \right)} \right) + \hat L
\end{equation}

In contrast, the H-F Branch divides the input features into two partitions: ${F_1} \in {\mathbb{R}^{N \times N \times \frac{C}{2}}}$ and ${F_2} \in {\mathbb{R}^{N \times N \times \frac{C}{2}}}$, to extract high-frequency information through a parallel architecture, respectively utilizing the sharp sensitivity of maximum filters and the detailed perception of Convs \cite{si2022inception}. $F_1$ successively passes through 1$\times$1 Conv layers and 3$\times$3 Conv layers to obtain ${\hat F_1}$. A combination of maximum pooling layers and 1x1 Conv layers is employed to appropriately compress the receptive field, yielding features ${\hat F_2}$. Finally, the concatenation of ${\hat F_1}$ and ${\hat F_2}$ generates a comprehensive feature map $H$ with rich high-frequency information: 
\begin{equation}
\label{eq_3}
{\hat F_1} = {{\mathop{\rm Conv}\nolimits} ^{3 \times 3}}\left( {{{{\mathop{\rm Conv}\nolimits} }^{1 \times 1}}\left( {{F_1}} \right)} \right)
\end{equation}

\begin{equation}
\label{eq_4}
{\hat F_2} = {{\mathop{\rm Conv}\nolimits} ^{1 \times 1}}\left( {{\mathop{\rm MaxPool}\nolimits} \left( {{F_2}} \right)} \right)
\end{equation}

\begin{equation}
\label{eq_5}
H = {\mathop{\rm Concat}\nolimits} \left( {{{\hat F}_1},{{\hat F}_2}} \right)
\end{equation}

The end of FIFB process involves the fusion of the low-frequency feature $L$ and the high-frequency feature $H$: 

\begin{equation}
\label{eq_6}
Z = L \oplus H
\end{equation}
where $\oplus$ represents element-wise addition of feature maps.

\begin{figure}
\centering
	\includegraphics[scale=.18]{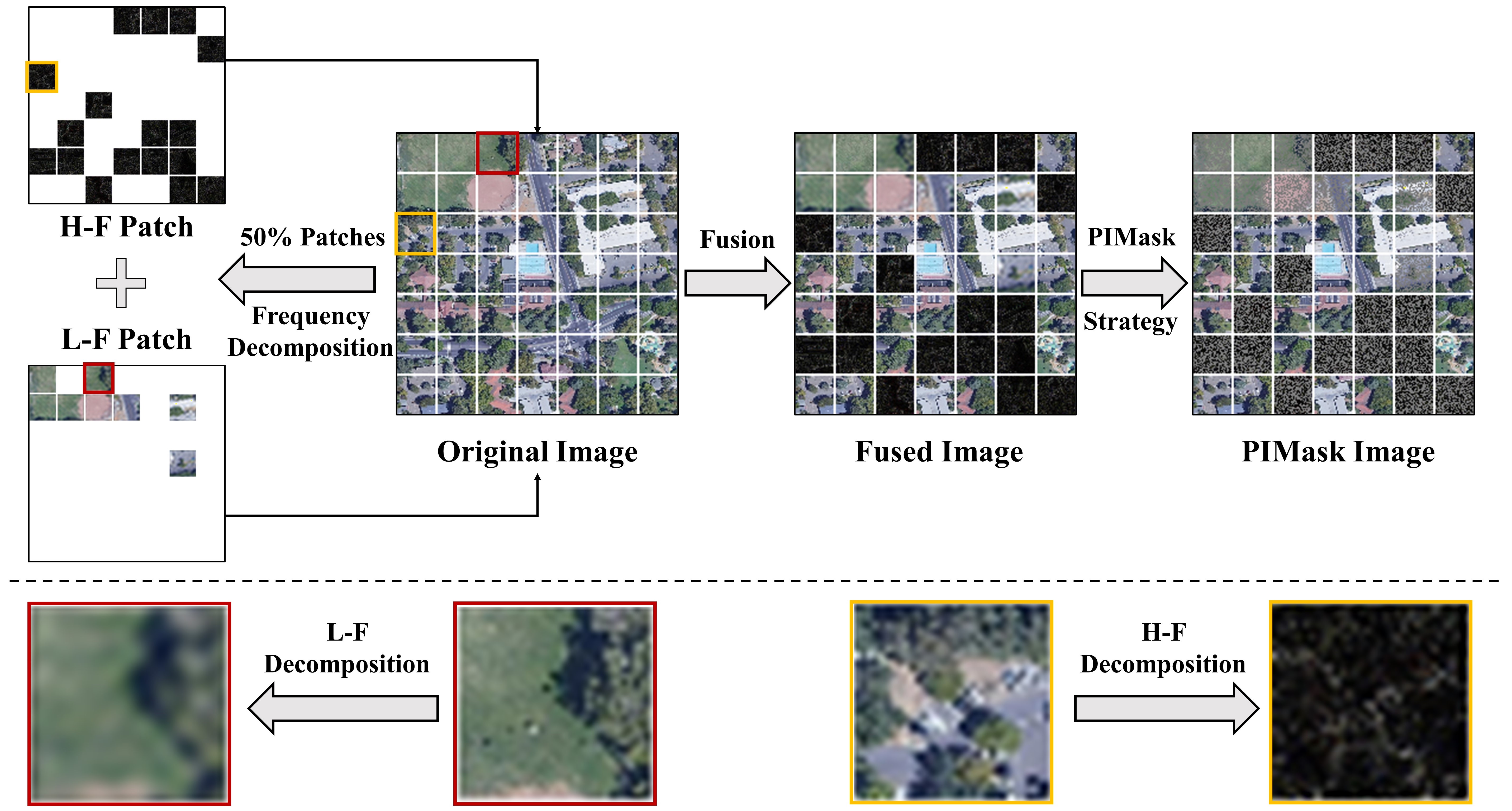}
\caption{ Illustration of FD-MIM Self-Supervised Pretraining method. First, 50\% of image patches are selected for frequency-domain analysis by algorithms like DFT (the above). Then, they are preserved critical frequency domain characteristics for decomposition (the below). Finally, random pixel masking (PIMask Strategy) is chosen to enhance the robustness and generalization.}
\label{fig_mim}
\end{figure}

\subsection{Frequency Domain Masked Image Modeling}
In computer vision tasks, it is a common practice to design a pretraining strategy that captures both local and global image features to enhance the efficiency and generalization ability of the model. One promising approach is using masking techniques to emphasize specific features in the images. Masked image modeling (MIM) \cite{he2022masked,xie2022simmim} can combine the intrinsic data relationship to guide the model to understand complex RS images better. By exploiting the structure of input images and the correlations among neighboring pixels, it enables the model to learn meaningful representations without explicit labeling.

Many MIM methods commonly employ the strategy of random masking. This technique involves selecting a certain proportion of image patches and subjecting them to complete masking. Although this method is widely used in natural images, it has challenges in the practical application of RS image interpretation. RS images have unique imaging mechanisms and contain more complex backgrounds and many smaller-scale objects, which limits many random masking strategies in RS image interpretation. In this context, we introduce the concept of high-frequency and low-frequency domain masked image modeling (FD-MIM). FD-MIM corresponds to the proposed CNN-Transformer hybrid framework.  Like other autoencoders, the proposed method can extract latent representations of masked images and use them to reconstruct the original signal of masked regions. By properly retaining the high-frequency and low-frequency domain information in the complex RS image, it contributes to better reconstruction details of the image while masking. The learned encoder is useful for various optical RS downstream tasks, and the L1 regression loss is used to calculate the difference between the reconstruction result and the pixel value.

The proposed FD-MIM strategy is shown in Fig. \ref{fig_mim}. Firstly, FD-MIM randomly selects 50\% of image patches from each RS image in the dataset. These patches are subjected to frequency-domain analysis, usually using techniques such as the Discrete Fourier Transform (DFT) to produce frequency-domain coefficients that describe the distribution of image energy over different spatial frequencies. Subsequently, the selected patches are classified into high-frequency or low-frequency categories. This classification depends on comparing the proportion of high-frequency content pixels to low-frequency content pixels within each patch. Patches with a higher proportion of high-frequency content are designated as high-frequency patches, while those dominated by low-frequency content are classified as low-frequency patches. To further emphasize high-frequency and low-frequency information, we perform high-pass and low-pass filtering on these classified patches, respectively. The former enhances the unique characteristics of the high-frequency portion, while the latter filtering contributes to preserving essential low-frequency information. This step facilitates a better separation of frequency components while preserving critical frequency domain characteristics. Finally, to enhance the robustness and generalization ability of the model, we introduce random pixel masking, which involves randomly selecting pixels from the frequency-separated patches and applying a masking operation. This strategy increases the complexity of the reconstruction images during training, facilitating the model to focus on learning the most relevant and discriminative features.

\section{Experiments}\label{exp}

\subsection{Remote Sensing Scene Classification}
\emph{1) Dataset Introduction:} We use three classic RS scene classification datasets to comprehensively evaluate the interpretation ability of the proposed lightweight foundation model.

\textbf{AID \cite{xia2017aid}.} This comprehensive dataset is curated from multi-sensor data collected via Google Earth's resources. The spatial resolution ranges from 8 meters to 0.5 meters, while the dimensions are set at 600 × 600 pixels. The AID dataset comprises a collection of 10,000 images, encompassing a total of 30 categories. The number of images within each category varies from 220 to 400 images. The AID dataset presents a combination of small inter-class variability, large intra-class variability, and an imbalanced distribution of categories, collectively posing a significant challenge to contemporary scene classification methods.

\textbf{NWPU-RESISC45 \cite{cheng2017remote}.} This dataset serves as a substantial benchmark sourced from Google Earth, presenting a broad spectrum of regions across the global landscape. It consists of 45 different categories, each of which contains 700 images, for a total of 31500 images. The spatial resolution of each RS image is noteworthy, ranging from approximately 0.2 to 30 meters, and standardized at a dimension of 256 × 256 pixels.

\textbf{UCM \cite{yang2010bag}.} The UCM dataset serves as a representative collection for scene classification, consisting of 2,100 RS images exclusively obtained from the USGS National Map. These images possess a uniform spatial resolution of 1 foot, maintaining a dimension of 256 pixels in both width and height. The assortment of images has been allocated across 21 distinct categories, amounting to a balanced distribution of 100 images per category. The dataset derives from diverse urban landscapes across the United States, thereby introducing a formidable level of complexity and challenge.

\emph{2) Detailed Experimental Settings:} All experiments are conducted using Docker images and PyTorch framework. The computational hardware employed is an NVIDIA Tesla A40 GPU, with a batch size set to 64. Training duration spans an average of 300 epochs. The model optimization employs the Adam optimizer until achieving convergence. The hyper-parameters are set to $\beta1=0.9$, $\beta2=0.999$ and $\epsilon=10^{-8}$. We load ImageNet pretrained weights for the ordinary method, while our method loads weights learned through MIM on 150000 visible images. The training sample proportions in the NWPU-RESISC dataset are configured as TR=10\% and TR=20\%, correspondingly. For the AID dataset, the training ratios are set at TR=20\% and TR=50\%. Meanwhile, the UCM dataset employs a training rate of TR=80\%. Each experiment is meticulously repeated five times to ensure robustness. Throughout these experiments, the metric utilized to evaluate performance is Overall Accuracy (OA), a widely adopted criterion in scene classification research.

\begin{table*}[h!t]
  \caption{Comparisons with the SOTA Models on Three Scene Classification Datasets under Different Settings.\\
  The OA Is Adopted as the Evaluation Metric.}
  \centering
  \label{tab1}
  \renewcommand\arraystretch{1.8}
  \setlength{\tabcolsep}{5.8mm}{
  \begin{tabular}{c|c|c|c|c|c}
  \toprule[1.5pt]

   \textbf{Method} & \textbf{\makecell[c]{AID \\ TR = 20\%}} & \textbf{\makecell[c]{AID \\ TR = 50\%}} &\textbf{\makecell[c]{NWPU-RESISC45 \\ TR = 10\%}} &\textbf{\makecell[c]{NWPU-RESISC45 \\ TR = 20\%}} & \textbf{\makecell[c]{UCM \\ TR = 80\%}}\\
   \midrule[0.75pt]
   MobileNet-v2 \cite{sandler2018mobilenetv2} & 73.82 &85.66 & 68.40 & 79.13 & 92.86\\
   MobileNet-v3s \cite{howard2019searching} & 75.15 & 82.54 & 63.39 & 74.92 & 91.90\\
   MobileNet-v3l \cite{howard2019searching} & 74.12 & 86.16 & 67.21 & 78.45 & 92.38\\
   ShuffleNet-v2-x1 \cite{ma2018shufflenet} & 72.29 & 85.10 & 64.57 & 75.81 & 91.19\\
   ShuffleNet-v2-x2 \cite{ma2018shufflenet} & 76.91 & 87.78 & 69.70 & 80.35 & 92.86\\
   EfficientNet-b0 \cite{tan2019efficientnet} & 75.06 & 85.72 & 69.23 & 80.50 & 92.62\\
   EfficientNet-b7 \cite{tan2019efficientnet} & 85.01 & 92.32 & 85.27 & 86.06 & 95.95\\
   MobileViT-xs \cite{mehta2021mobilevit} & 78.94 & 88.94 & 71.44 & 80.66 & 94.76\\
   MobileViT-s \cite{mehta2021mobilevit} & 79.29 & 88.82 & 71.89 & 78.98 & 95.48\\
   ResNet-50 \cite{he2016deep} & 77.34 & 88.06 & 68.51 & 80.75 & 93.57\\
   Swin Tiny (MIM) \cite{liu2021swin} & 89.04 & 95.18 & 86.67 & 91.62 & 95.71\\
   \midrule[0.75pt]
   \textbf{RingMo-lite} & \textbf{93.86} &\textbf{96.54} & \textbf{89.85}& \textbf{93.25} & \textbf{99.05}\\
  
  \bottomrule[1.5pt]
  \end{tabular}}
\end{table*}

\begin{table}[t]
  \caption{Comparisons of Params and FLOPs of Different Methods on UCM Dataset.}
  \centering
  \label{tab2}
  \renewcommand\arraystretch{1.5}
  \setlength{\tabcolsep}{2.6mm}{
  \begin{tabular}{c|c|c|c}
  \toprule[1.5pt]

   \textbf{Method} & \textbf{Para. (M)} & \textbf{FLOPs (G)} & \textbf{OA (\%)} \\
   \midrule[0.75pt]
   ResNet-50 \cite{he2016deep} & 23.551 & 4.132 & 93.57\\
   Swin Tiny (Supervised) & 28.288 & 4.494 & 92.86\\
   Swin Tiny (FIFB) & 28.294 & 4.494 & 93.10\\
   Swin Tiny (MIM) & 28.288 & 4.494 & 95.71\\
   RingMo-lite & 28.294 & 4.494 & \textbf{99.05}\\
   \midrule[0.75pt]
   RingMo \cite{sun2022ringmo} & 87.768 & 15.438 & 99.06\\
  
  \bottomrule[1.5pt]
  \end{tabular}}
\end{table}

\begin{figure}[ht]
\centering
	\includegraphics[scale=.35]{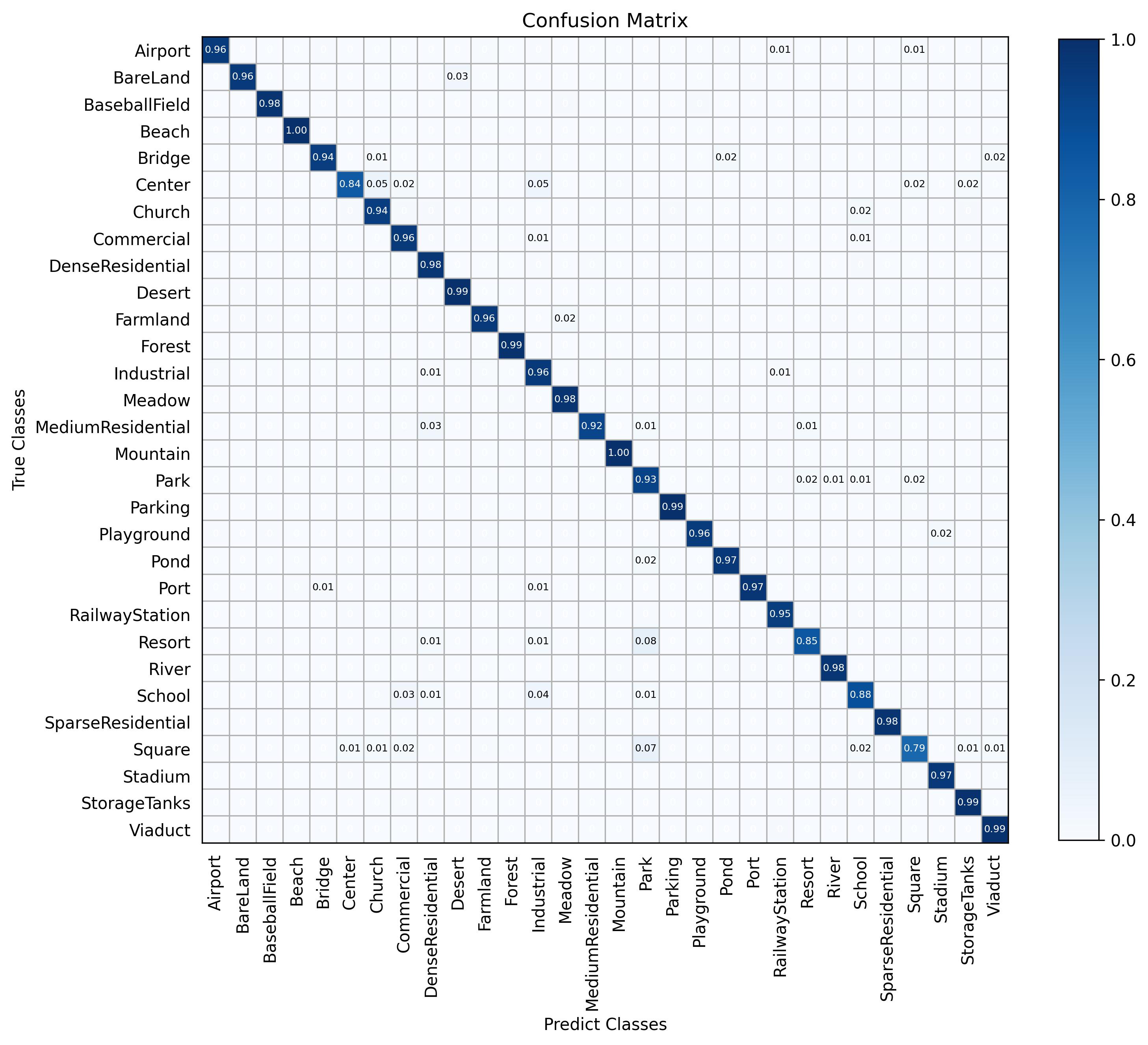}
\caption{Classification confusion matrix for RingMo-lite on 50\% AID training data.}
\label{fig_AID}
\end{figure}

\begin{figure}[ht]
\centering
	\includegraphics[scale=.35]{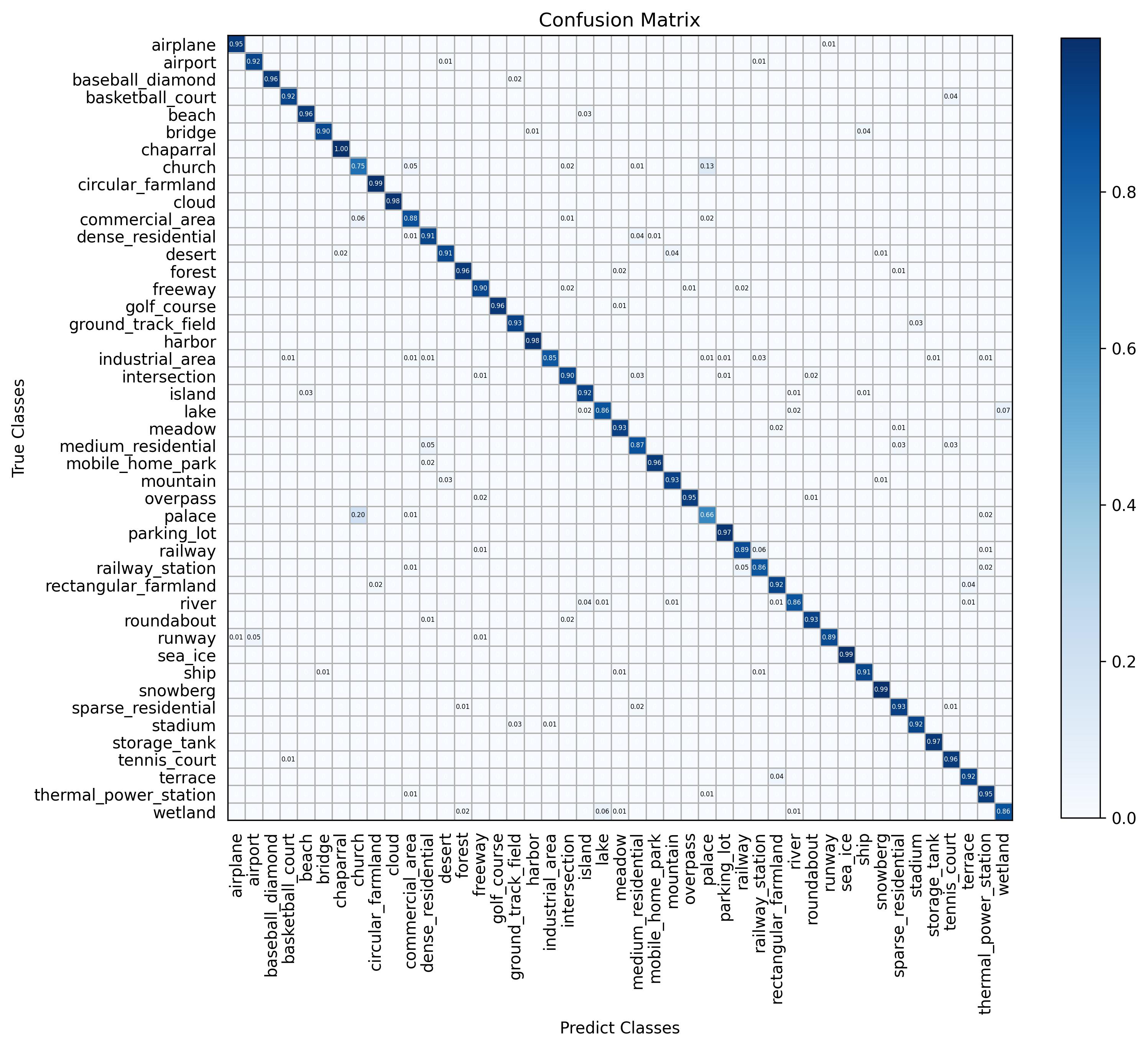}
\caption{Classification confusion matrix for RingMo-lite on 20\% NWPU-RESISC45 training data.}
\label{fig_NWPU}
\end{figure}

\begin{figure}[ht]
\centering
	\includegraphics[scale=.35]{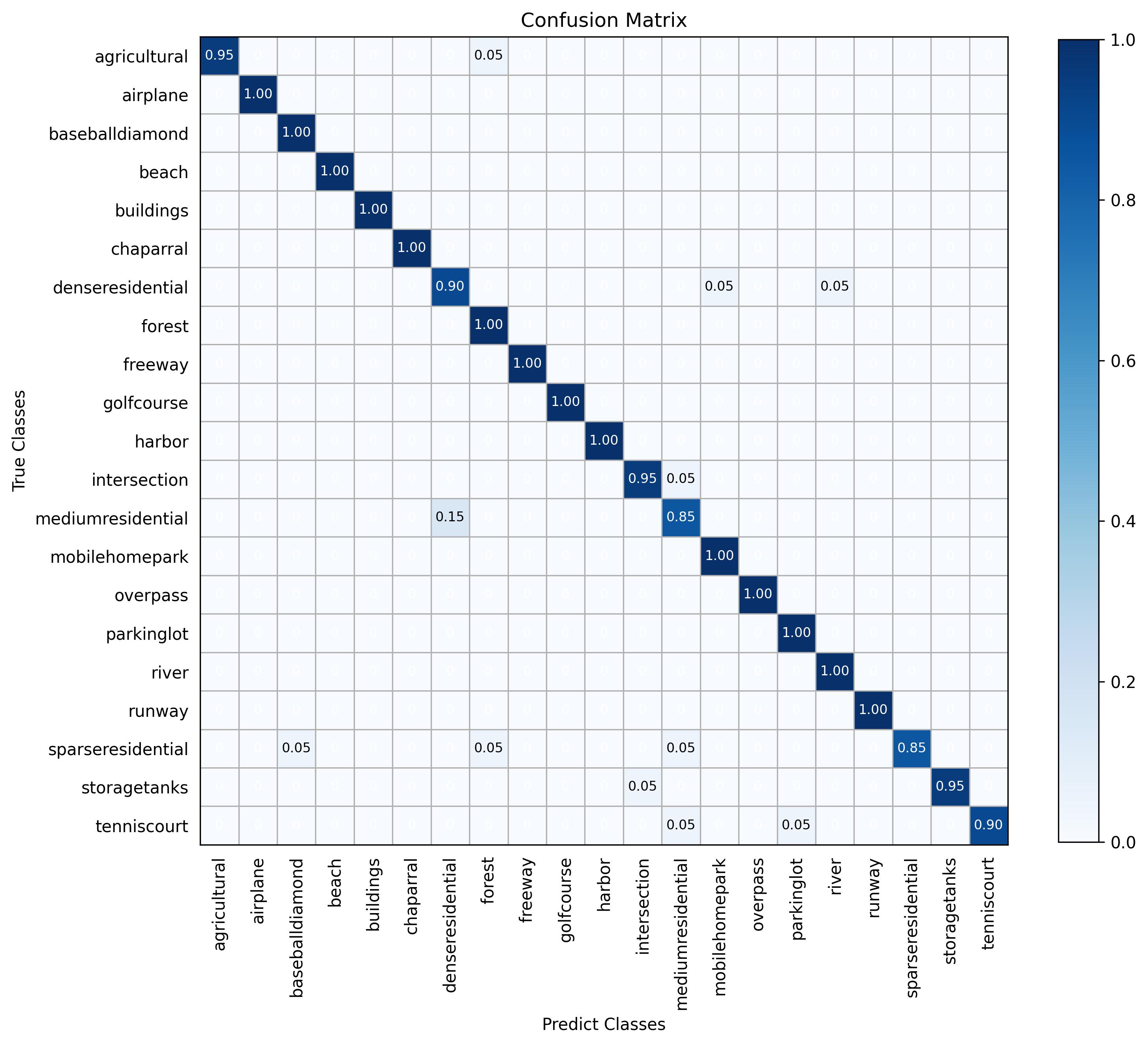}
\caption{Classification confusion matrix for RingMo-lite on 80\% UCM training data.}
\label{fig_UCM}
\end{figure}

\emph{3) Experimental Result Analysis:} Table \ref{tab1} shows the results of our method on different datasets, compared with the conventional lightweight algorithms mentioned above. "Swin Tiny (MIM)" and "RingMo-lite" represent the baseline of our proposed method and the version after adding the FIFB module, respectively. Obviously, compared with conventional lightweight methods, our method has achieved the best results on various datasets. Especially in the scene classification tasks of the difficult AID and NWPU datasets, the performance of the Swin-based method has been significantly improved, which demonstrates the superiority of our approach. To further illustrate the significant effect of the proposed modules, a comparison with the baseline is performed, and the results show that the introduction of the FIFB module improves the OA by 1.36\%-5.48\% on different datasets and training sample proportions, indicating that the proposed module has a good feature extraction effect and high robustness on scene classification tasks.

In order to prove the superiority of our method from different perspectives, we also compare the parameters and computation size of different methods, and the results are presented in Table \ref{tab2}. It should be noted that we use the floating point operations (FLOPs) to represent the amount of model calculations, and the results of OA are obtained based on the UCM dataset. Through comprehensive analysis of the results, we can draw the following conclusions. First, compared with the ResNet-50 method \cite{he2016deep} with the same parameter level, our method improves the OA by 5.48\%. Secondly, the accuracy of the RingMo method \cite{sun2022ringmo} in the table is obtained by pretraining 100 epochs on 1.5 million data and then fine-tuning. In comparison, our method has one-third the number of parameters and one-fourth the computation. With such a large gap, the OA of our method is only 0.01\% lower than it. Finally, the MIM self-supervised pretraining does not change the size of the original network parameters, while the introduction of the FIFB module only adds 0.006M parameters. However, compared with the supervised Swin Tiny method and the MIM self-supervised pretrained Swin Tiny method, our method improves the OA by 6.19\% and 3.34\%, respectively. In summary, our method maximizes the effect of RingMo while significantly optimizing the amount of parameters and calculations.

Fig. \ref{fig_AID}-\ref{fig_UCM} are the classification confusion matrices on the AID, NWPU, and UCM datasets, showing the prediction results of our proposed RingMo-lite method among different categories. Experimental results show that our method achieves excellent performance on almost all scene categories.

\subsection{Remote Sensing Object Detection}

\begin{table*}[h!t]
  \caption{Comparisons with the SOTA Models on DIOR Datasets.}
  \centering
  \label{tab3}
  \renewcommand\arraystretch{1.8}
  \setlength{\tabcolsep}{3.8mm}{
  \begin{tabular}{c|c|c|c|c|c|c|c|c|c|c|c}
  \toprule[1.5pt]

   \multirow{2}{*}{\textbf{Method}} & \textbf{C1} & \textbf{C2} & \textbf{C3} & \textbf{C4} & \textbf{C5} & \textbf{C6} & \textbf{C7} & \textbf{C8} & \textbf{C9} & \textbf{C10} &\multirow{2}{*}{\textbf{mAP}} \\
   
   \multirow{2}{*}{} & \textbf{C11} & \textbf{C12} & \textbf{C13} & \textbf{C14} & \textbf{C15} & \textbf{C16} & \textbf{C17} & \textbf{C18} & \textbf{C19} & \textbf{C20} & \multirow{2}{*}{}\\
   \midrule[0.75pt]
   
   \multirow{2}{*}{FasterRCNN (r18) \cite{ren2015faster}} & 46.4 & 43.9 &62.4 & 66.3 & 24.4 & 62.7 & 45.8& 60.1 & 44.5 & 50.8 &\multirow{2}{*}{50.9} \\
   
   \multirow{2}{*}{} & 57.4 & 48.1 &44.0 & 60.9 & 54.0 & 34.6 & 77.2& 35.5 & 27.3 & 70.8 & \multirow{2}{*}{}\\
   \midrule[0.50pt]
   
   \multirow{2}{*}{RetinaNet (r18) \cite{lin2017focal}} & 50.5 & 58.9 &64.3 & 75.0 & 26.6 & 66.7 & 52.7& 68.5 & 49.1 & 61.9 &\multirow{2}{*}{55.6} \\
   
   \multirow{2}{*}{} & 64.6 & 51.4 &47.6 & 58.1 & 69.4 & 33.8 & 77.7& 33.7 & 27.8 & 73.2 & \multirow{2}{*}{}\\
   \midrule[0.50pt]
   
   \multirow{2}{*}{Reppoints (r18) \cite{yang2019reppoints}} & 55.8 & 61.4 &67.5 & 74.0 & 31.7 & 68.2 & 51.5& 68.1 & 49.2 & 56.0 &\multirow{2}{*}{57.9} \\
   
   \multirow{2}{*}{} & 66.4 & 56.3 &50.1 & 65.3 & 69.8 & 43.9 & 78.4& 41.4 & 30.1 & 73.9 & \multirow{2}{*}{}\\
   \midrule[0.50pt]
   
   \multirow{2}{*}{FasterRCNN (r50) \cite{ren2015faster}} & 54.1 & 75.9 &65.8 & 84.8 & 35.7 & 77.6& 61.8& 74.7 & 54.3 & 77.7 &\multirow{2}{*}{63.7} \\
   
   \multirow{2}{*}{} & 77.2 & 77.2 &57.1 & 64.5 & 63.5 & 43.9 & 83.6& 56.2 & 31.3 & 79.9 & \multirow{2}{*}{}\\
   \midrule[0.50pt]
   
   \multirow{2}{*}{RetinaNet (r50) \cite{lin2017focal}} & 62.0 & 79.2 &72.2 & 86.7 & 34.4 & 79.5& 63.2& 76.6 & 55.0 & 81.3 &\multirow{2}{*}{64.5} \\
   
   \multirow{2}{*}{} & 75.8 & 48.8 &55.7 & 59.2 & 64.4 & 42.4 & 84.6& 53.0 & 35.1 & 81.1 & \multirow{2}{*}{}\\
   \midrule[0.50pt]
   
   \multirow{2}{*}{Reppoints (r50) \cite{yang2019reppoints}} & 60.8 & 83.7 &75.5 & 87.5 & 43.5 & 76.8& 67.1& 79.8 & 60.5 & 81.2 &\multirow{2}{*}{69.8} \\
   
   \multirow{2}{*}{} & 78.8 & 58.2 &59.7 & 73.3 & 71.7 & 60.7 & 86.6& 62.7 & 43.9 & 84.8 & \multirow{2}{*}{}\\
   \midrule[0.50pt]
   
   \multirow{2}{*}{Reppoints (Swin-Tiny)} & 60.3 & 89.3 & 78.0 & 86.5 & 42.7 &77.5 &71.9 & 86.0 & 67.0 & 83.0 &\multirow{2}{*}{71.9} \\
   
   \multirow{2}{*}{} & 80.7 & 60.1 &60.5 & \textbf{72.6} & \textbf{77.4} &60.7 & 86.9& \textbf{69.2} & 42.2 & 84.6 & \multirow{2}{*}{}\\
   \midrule[0.50pt]
   
   \multirow{2}{*}{\textbf{RingMo-lite}} & \textbf{64.8} & \textbf{90.5} &\textbf{81.0} & \textbf{87.6} & \textbf{44.1} &\textbf{79.9} &\textbf{76.8} & \textbf{86.6} & \textbf{67.3} & \textbf{85.0} &\multirow{2}{*}{\textbf{73.4}} \\
   
   \multirow{2}{*}{} & \textbf{82.5} & \textbf{60.9} & \textbf{61.8}& 72.3 & 77.0 & \textbf{62.5} &\textbf{87.7} & 68.9 & \textbf{44.2} & \textbf{86.3} & \multirow{2}{*}{}\\
   
   

  \bottomrule[1.5pt]
  \end{tabular}}
  \begin{threeparttable}
  \begin{tablenotes}
    \footnotesize
    \item[*] The correspondence between the abbreviations in the table and the categories. C1: Airplane, C2: Airport, C3: Baseball Field, C4: Basketball Court, C5: Bridge, C6: Chimney, C7: Dam, C8: Expressway Service Area, C9: Expressway Toll Station, C10: Golf Course, C11: Ground Track Field, C12: Harbor, C13: Overpass, C14: Ship, C15: Stadium, C16: Storage Tank, C17: Tennis Court, C18: Train Station, C19: Vehicle, C20: Wind Mill
  \end{tablenotes}
 \end{threeparttable}
\end{table*}

\begin{table}[h!t]
  \caption{Comparisons of Params and FLOPs of Different Methods on DIOR Dataset.}
  \centering
  \label{tab4}
  \renewcommand\arraystretch{2.3}
  \setlength{\tabcolsep}{2.7mm}{
  \begin{tabular}{c|c|c|c}
  \toprule[1.5pt]

   \textbf{Method} & \textbf{Para. (M)} & \textbf{FLOPs (G)} & \textbf{mAP (\%)} \\
   \midrule[0.75pt]
   RetinaNet (r50) \cite{lin2017focal} & 36.5 & 212.77 & 64.5\\
   Reppoints (r50) \cite{yang2019reppoints} & 36.6 & 189.87 & 69.8\\
   \makecell[c]{Reppoints\\(Swin Tiny)(Supervised)} & 37.32& 195.36 & 71.9\\
   \makecell[c]{Reppoints\\(Swin Tiny)(MIM)} & 37.32 & 195.36 & 72.7\\
   \makecell[c]{RingMo-lite} & 37.36 & 196.07 & \textbf{73.4}\\
   \midrule[0.75pt]
   RingMo \cite{sun2022ringmo} & 97.14 & 450.66 & 74.7\\
  \bottomrule[1.5pt]
  \end{tabular}}
\end{table}

\begin{table*}[h!t]
  \caption{Comparisons with the SOTA Models on FAIR-1M Datasets.}
  \centering
  \label{tab_det_fair1m}
  \renewcommand\arraystretch{1.8}
  \setlength{\tabcolsep}{1.5mm}{
  \begin{tabular}{c|c|c|c|c|c|c|c|c|c|c|c|c|c|c|c|c|c|c}
  \toprule[1.5pt]

   \multirow{2}{*}{\textbf{Method}} &  \textbf{C1} & \textbf{C2} & \textbf{C3} & \textbf{C4} & \textbf{C5} & \textbf{C6} & \textbf{C7} & \textbf{C8} & \textbf{C9} & \textbf{C10} & \textbf{C11} & \textbf{C12} & \textbf{C13} & \textbf{C14} & \textbf{C15} & \textbf{C16} & \textbf{C17} &\multirow{2}{*}{\textbf{mAP}} \\
   
   \multirow{2}{*}{} & \textbf{C18} & \textbf{C19} & \textbf{C20} & \textbf{C21} & \textbf{C22} & \textbf{C23} & \textbf{C24} & \textbf{C25} & \textbf{C26} & \textbf{C27} & \textbf{C28} & \textbf{C29} & \textbf{C30} & \textbf{C31} & \textbf{C32} & \textbf{C33} & \textbf{C34} & \multirow{2}{*}{}\\
   \midrule[0.75pt]
   
   \multirow{2}{*}{RetinaNet \cite{lin2017focal}} & 38.5 & 55.4 & 24.8 & 51.8 & 0.8 & 40.5 & 41.1 & 18.0 & 19.9 & 1.7 & 9.6 & 22.6 & 1.3 & 16.4 & 19.1 & 14.3 & 24.7  &\multirow{2}{*}{30.7} \\
   
   \multirow{2}{*}{} & 15.4 & 65.2 &22.4 & 44.2 & 35.4 & 52.4 & 19.2 & 1.3 & 17.0 & 29.0 & 50.6 & 81.1 & 52.5 & 66.8 & \textbf{60.1} & 17.4 & 12.6 & \multirow{2}{*}{}\\
   \midrule[0.50pt]
   
   \multirow{2}{*}{Cascade-RCNN \cite{cai2018cascade}} & 40.4 & 52.9 & \textbf{29.1} & 52.5  & 0 & 44.4 & 38.4 & 26.6 & 17.5 & 0 & 12.1 & 28.8 & 0.7 & 15.4 & 18.5 & 14.6 & 25.2 &\multirow{2}{*}{ 31.2 } \\
   
   \multirow{2}{*}{} & 14.5 & 68.2 & 28.3 & 48.6 & 40.4 & 58 & 13.7 & 0.9 & 16.5 & 30.3 & 38.8 & 80.3 & 48.2 & 67.9 & 55.7 & 20.4 & 12.6 & \multirow{2}{*}{}\\
   \midrule[0.50pt]
   
   \multirow{2}{*}{FasterRCNN \cite{ren2015faster}} & 31.9 & 81.0 &15.1 & 48.1 & 0 & 45.3 & \textbf{50.9}& 52.8 & 46.9 & 50.9 & 9.0 & 32.7 & 3.4 & 20.3 &15.2 & \textbf{66.6} & \textbf{68.1} &\multirow{2}{*}{32.7} \\
   
   \multirow{2}{*}{} & 20.4 & 47.8 &0 & 23.8 & 21.0 & 43.6 & 0& 0 & 0 & 0 & 30.3 & 40.1 & 56.1 & \textbf{87.5} &51.5 & \textbf{31.2} & 9.2 & \multirow{2}{*}{}\\
   \midrule[0.50pt]

    \multirow{2}{*}{RoI Transformer \cite{ding2019learning}} & 39.6 & 73.6 & 18.3 & \textbf{56.4} & 0 & 47.7 & 49.9 & 27.6 & 31.8 & 0 & 14.3 & 28.1 & 1.0 & 14.3 & 16.0 & 18.0 & 26.0 &\multirow{2}{*}{35.3 } \\
   
   \multirow{2}{*}{} & 13.0 & \textbf{68.8} & \textbf{37.4} & 54.0 & 45.7 & 58.4 & 16.2 & \textbf{5.1} & \textbf{22.2} & \textbf{46.7} & \textbf{54.8} & 80.4 & \textbf{56.7} & 69.1 & 58.4 & 18.6 & 31.8 & \multirow{2}{*}{}\\
   \midrule[0.50pt]
   
   \multirow{2}{*}{FCOS \cite{tian2019fcos}} & 38.7 & 82.7 & 23.6 & 55.6 & 25.5 & 47.8 & 64.7 & \textbf{57.3} & \textbf{64.6} & 32.9 & 16.6 & 28.1 & 11.7 & \textbf{29.5} & 15.7 & 26.8 & 39.0 &\multirow{2}{*}{ 42.5 } \\
   
   \multirow{2}{*}{} & 33.8 & 61.2 & 30.0 & 49.0 & 51.3 & \textbf{58.4} & 16.3 & 3.0 & 17.0 & 5.4 & 49.2 & \textbf{81.1} & 52.5 & 67.0 & 54.6 & 25.7 & \textbf{33.1} & \multirow{2}{*}{}\\
   \midrule[0.50pt]
   
   \multirow{2}{*}{\textbf{RingMo-lite}} & \textbf{43.2} & \textbf{85.0} & 26.0 & 53.5 & \textbf{45.4} & \textbf{53.8} & \textbf{67.8} & 56.1 & 63.6 & 33.8 & \textbf{16.6} & \textbf{34.4} & \textbf{13.7} & 26.3 & \textbf{21.4} & 42.2 & 42.7 &\multirow{2}{*}{ \textbf{44.1} } \\
   
   \multirow{2}{*}{} & \textbf{34.3} & 62.0 & 33.3 & \textbf{54.0} & \textbf{51.4} & 57.9 & \textbf{20.2} & 4.9 & 17.2 & 7.7 & 46.7 & 80.4 & 53.6 & 68.8 & 55.9 & 28.6 & 31.8 & \multirow{2}{*}{}\\

  \bottomrule[1.5pt]
  \end{tabular}}
  \begin{threeparttable}
  \begin{tablenotes}
    \footnotesize
    \item[*] The correspondence between C1-C34 and object categories in the table (in order). C1: Boeing 737, C2: Boeing 747, C3: Boeing 777, C4: Boeing 787, C5: C919, C6: A220, C7: A321, C8: A330, C9: A350, C10: ARJ21, C11: passenger ship, C12: motorboat, C13: fishing boat, C14: tugboat, C15: engineering ship, C16: liquid cargo ship, C17: dry cargo ship, C18: warship, C19: small car, C20: bus, C21: cargo truck, C22: dump truck, C23: van, C24: trailer, C25: tractor, C26: excavator, C27: truck tractor, C28: basketball court, C29: tennis court, C30: football field, C31: baseball field, C32: intersection, C33: roundabout, C34: bridge
  \end{tablenotes}
 \end{threeparttable}
\end{table*}

\begin{table}[h!t]
  \caption{Comparisons of Params and FLOPs of Different Methods on FAIR1M Dataset.}
  \centering
  \label{tab_fair1m}
  \renewcommand\arraystretch{2.3}
  \setlength{\tabcolsep}{1.5mm}{
  \begin{tabular}{c|c|c|c}
  \toprule[1.5pt]

   \textbf{Method} & \textbf{Para. (M)} & \textbf{FLOPs (G)} & \textbf{mAP (\%)} \\
   \midrule[0.75pt]
   RoI Transformer (r50) \cite{ding2019learning}  & 55.29 & 225.44 & 35.3\\
   FCOS  (r50) \cite{tian2019fcos} & 31.97 & 208.01 & 42.5\\
   \makecell[c]{FCOS(Swin Tiny)(Supervised)} & 35.07 & 215.7 & 42.8\\
   \makecell[c]{FCOS(Swin Tiny)(MIM)} & 35.07 & 215.7 & 43.8\\
   \makecell[c]{RingMo-lite} & 35.11 & 215.7 & \textbf{44.1}\\
   \midrule[0.75pt]
   RingMo \cite{sun2022ringmo} & 104.13 & 455.51 & 45.8\\
  \bottomrule[1.5pt]
  \end{tabular}}
\end{table}

\emph{1) Dataset Introduction:} We validate the performance of our model on the DIOR dataset covering complex component objects (horizontal object detection task) and the RS fine-grained category dataset FAIR1M (oriented object detection task).

\textbf{DIOR \cite{li2020object}.} DIOR stands as an expansive and publicly accessible RS dataset that includes 20 distinct object categories. The dataset includes 23463 images, totaling 190288 instances. Each image is meticulously annotated with horizontal bounding boxes, which facilitate both model training and testing phases. Beyond its diverse array of RS scenes, DIOR also encapsulates objects of different geometric shapes. Notably, within the context of RS, some objects have a wide coverage range, often composed of multiple regular components. Examples of these complex composite objects within DIOR include infrastructures like airports, expressway service areas, Golf courses, and train stations, each posing a heightened level of challenge to detection algorithms. Our experimental endeavors place particular emphasis on delving into the performance outcomes of these complex composite object categories.

\textbf{FAIR1M \cite{sun2022fair1m}.} The dataset encompasses over 1 million instances. It comprises a repository of more than 15000 images, captured across various platforms, boasting resolutions spanning from 0.3m to 0.8m. The whole objects are categorized into 5 primary classes and subsequently further subdivided into 37 intricate sub-categories by oriented bounding boxes. These primary classes encompass distinct entities such as airplanes, vehicles, ships, roads, and courts, with each of them comprising a diverse array of fine-grained sub-categories. To streamline the experimental framework, we exclude three other categories, resulting in a focused assessment of the remaining 34 model categories. For the sake of clarity and coherence throughout our experiments, we assign the aforementioned categories with designations ranging from C1 to C34.

\emph{2) Detailed Experimental Settings:} The GPU and Docker image are consistent to the prior classification experiments. Training parameters are tuned to encourage model convergence, with a learning rate of 0.0005, momentum at 0.99 and weight decay at 0.0001. We evaluate our models on the test part of the DIOR and FAIR1M datasets, following the open-source project of MMDetection and MMRotate, respectively.

\begin{figure*}
\centering
	\includegraphics[scale=0.14]{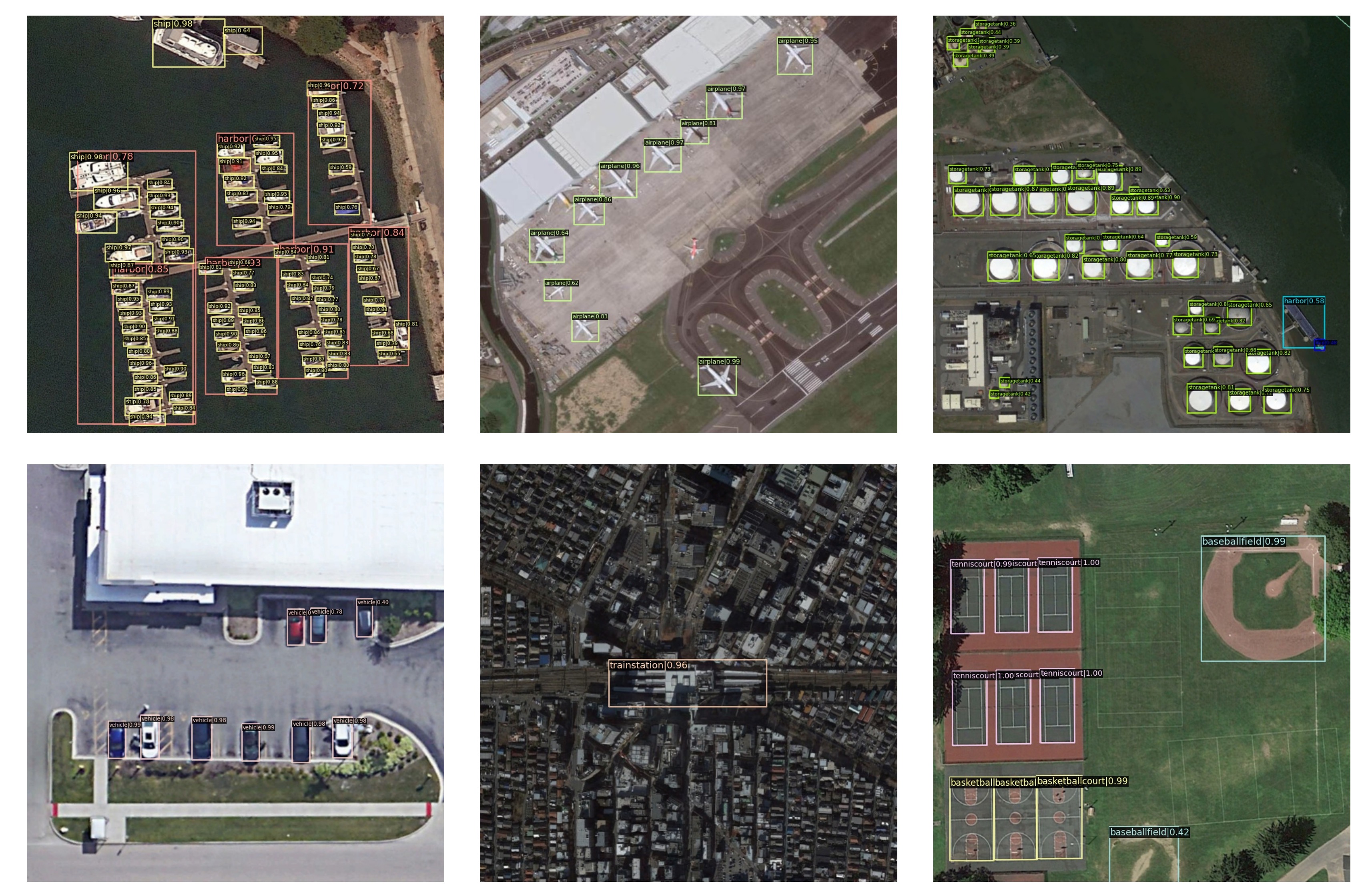}
\caption{Visualization results of the proposed RingMo-lite on the DIOR dataset.}
\label{fig_dior}
\end{figure*}

\begin{figure*}
\centering
	\includegraphics[scale=0.12]{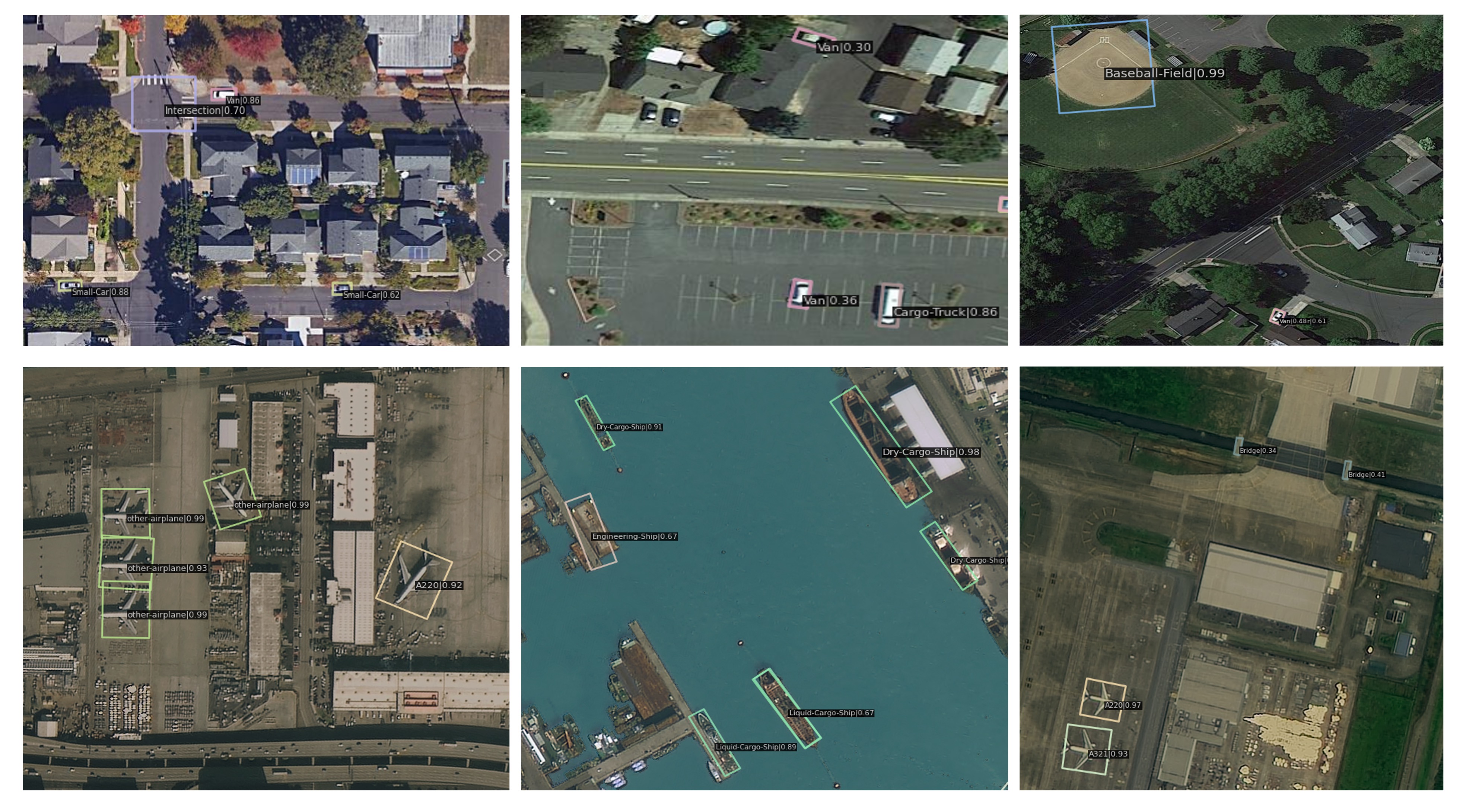}
\caption{Visualization results of the proposed RingMo-lite on the FAIR1M dataset.}
\label{fig_fair}
\end{figure*}

\emph{3) Experimental Result Analysis:} Table \ref{tab3} shows the Horizontal Bounding Box (HBB) detection accuracy of various methods on the DIOR dataset, which can be simply divided into three categories: methods based on ResNet-18 \cite{he2016deep}, ResNet-50 \cite{he2016deep} and Swin Tiny \cite{liu2021swin} as the backbone. FasterRCNN \cite{ren2015faster}, RetinaNet \cite{lin2017focal}, and Reppoints \cite{yang2019reppoints} represent different object detection algorithms, which will be combined with two kind of ResNet-based backbones. "Reppoints (Swin-Tiny)" means the combination of Swin Tiny backbone with the Reppoints detection head. Through the analysis of the experimental results, the method based on Reppoints has the best effect among the three methods. In addition, the proposed RingMo-lite improved the detection accuracy by 1.5\%. The Reppoints uses corner point regression to calculate the location of key points, which may have a better representation of RS objects.

Table \ref{tab_det_fair1m} shows the Oriented Bounding Box (OBB) detection accuracy of various methods on the FAIR1M dataset, which can be simply divided into two parts: methods based on CNNs like RetinaNet \cite{lin2017focal}, Cascade-RCNN \cite{cai2018cascade}, FasterRCNN \cite{ren2015faster} and FCOS \cite{tian2019fcos}. On the other hand, RoI Transformer \cite{ding2019learning} is based on Transformer. Through the analysis of the experimental results, we find that the method of 
FCOS achieves the best accuracy. Meanwhile, RingMo-lite, which adds FIFB module and MIM, improves the detection accuracy by 1.6\%, which shows our method has an advance in rotated object detection. 

At the same time, we also compare the parameters and calculation amount of different methods, and the results are presented in Table \ref{tab4} and \ref{tab_fair1m}.  We still measure the amount of calculation through FLOPs, and the results of mAP on both of the DIOR  and FAIR1M dataset. It can be seen that for HBB, the parameters of the RingMo-lite are  reduced to one-third of the RingMo foundation model, while the detection accuracy mAP is only reduced by 1.3\%. For OBB, the parameters are decreased to 33\% and the accuracy is 1.7\%. Secondly, for HBB, compared to the baselines of ResNet-50 and Swin Tiny, the parameters of RingMo-lite have increased slightly by 0.76M and 0.04M, respectively, but the effect has improved by 3.6\% and 0.7\%. For OBB, the increasing parameters of Swin Tiny is 3.1M and the mAP improvement is 1.6\% and 0.3\%, respectively. Finally, for HBB, although the number of parameters is equivalent, the calculation of RingMo-lite is reduced by 7.5\% compared with RetinaNet (r50). For OBB, FCOS has 40\% fewer parameters compared with RoI Transformer (r50).

We show the visualization results on some of the samples in DIOR and FAIR1M in Fig. \ref{fig_dior} and \ref{fig_fair}. It can be seen that RingMo-lite has a good detection effect on common RS targets, from small cars to large buildings, airplanes, ships and infrastructure.

\begin{table*}[h!t]
  \caption{Comparisons with the SOTA Semantic Segmentation Methods on the Validation Set of iSAID.}
  \centering
  \label{tab5}
  \renewcommand\arraystretch{1.8}
  \setlength{\tabcolsep}{1.0mm}{
  \begin{tabular}{c|c|c|c|c|c|c|c|c|c|c|c|c|c|c|c|c}
  \toprule[1.5pt]
  
   \multirow{2}{*}{\textbf{Method}} & \multicolumn{15}{c|}{\textbf{Acc per category(\%)}} &\multirow{2}{*}{\textbf{mIoU(\%)}} \\
   \cline{2-16}
   \multirow{2}{*}{} & \textbf{BD} & \textbf{BC} & \textbf{Bridge} & \textbf{GTF} & \textbf{Harbor} & \textbf{HC} & \textbf{LV} & \textbf{Plane} & \textbf{RA} & \textbf{Ship} & \textbf{SV} & \textbf{SBF} & \textbf{ST} & \textbf{SP} & \textbf{TC} & \multirow{2}{*}{}\\
   
   \midrule[0.75pt]
   
   Semantic FPN \cite{kirillov2019panoptic} & 71.82 & 57.88 & 33.96 & 51.60 & 51.33 & 0 & 59.19 & 80.84 & \textbf{68.66} & 63.67 & 45.08 & 73.62 & 59.51 & 46.35 & 86.62 & 59.32\\
   
   PSPNet (r18) \cite{zhao2017pyramid} & 75.25 & 57.85 & 38.21 & 54.25 & 52.08 & 0 & 59.65 & 80.90 & 60.73 & 64.30 & 43.46 & 73.79 & 69.77 & 47.09 & 87.21 & 60.22\\
   
   DeeplabV3+ (r18) \cite{chen2018encoder} & 73.12 & 59.78 & 37.46 & 56.19 & 56.91 & 0 & 61.91 & 82.13 & 63.11 & 66.24 & 46.72 & 73.84 & 71.48 & 46.55 & 87.19 & 61.35\\
   
   PSPNet (r50) \cite{zhao2017pyramid} & 77.87 & \textbf{65.78} & 41.50 & 54.85 & 56.30 & 35.44 & 63.38 & 83.96 & 65.54 & 68.03 & 48.88 & 75.11 & \textbf{74.11} & 47.28 & 88.67 & 65.36\\

   DeeplabV3+ (r50) \cite{chen2018encoder} & 76.40 & 60.07 & 43.90 & 58.88 & \textbf{58.17} & 33.33 & 64.88 & \textbf{84.97} & 67.48 & \textbf{70.93} & 51.77 & \textbf{76.92} & 73.74 & \textbf{50.40} & \textbf{88.88} & 66.24\\
   
   DeeplabV3+ (Swin Tiny) \cite{chen2018encoder} & \textbf{77.96} & 62.06 & 42.17 & \textbf{60.63} & 55.75  & 38.38 & 65.03 & 84.01 & 69.0 & 69.12 & 49.02 & 76.35 & 73.34 & 49.86 & 88.44& 66.27\\
   \makecell[c]{RingMo-lite} & 76.92 & 65.03 & \textbf{43.94} & 59.4 & 55.69 & \textbf{42.26} & \textbf{65.96} & 84.1 & 68.1 & 69.94 & \textbf{49.43} & 75.68 & 72.44 & 48.73 & 87.69 & \textbf{66.53} \\
   
  \bottomrule[1.5pt]
  \end{tabular}}
  \begin{threeparttable}
  \begin{tablenotes}
    \footnotesize
    \item[*] The 15 foreground categories of iSAID dataset: baseball diamond (BD), basketball court (BC) bridge, ground track field (GTF), harbor, helicopter (HC), large vehicle (LV), plane, roundabout (RA), ship, small vehicle (SV), soccer ball field (SBF), storage tank (ST), swimming pool (SP) and tennis court (TC)
  \end{tablenotes}
 \end{threeparttable}
  
\end{table*}

\begin{table}[h!t]
  \caption{Comparisons of Params and FLOPs of Different Methods on iSAID Dataset.}
  \centering
  \label{tab_isaid_para}
  \renewcommand\arraystretch{2.3}
  \setlength{\tabcolsep}{2.7mm}{
  \begin{tabular}{c|c|c|c}
  \toprule[1.5pt]

   \textbf{Method} & \textbf{Para. (M)} & \textbf{FLOPs (G)} & \textbf{OA (\%)} \\
   \midrule[0.75pt]
   DeeplabV3+ (r50) \cite{chen2018encoder}  & 41.219 & 177 & 66.24\\
   \makecell[c]{DeeplabV3+\\ (Swin Tiny)(Supervised)}  & 41.885 & 38.419 & 66.27\\
   \makecell[c]{DeeplabV3+\\(Swin Tiny)(MIM)} & 41.885 & 38.419 & 66.41\\
   \makecell[c]{RingMo-lite} & 41.912 & 38.746 & \textbf{66.53}\\
   \midrule[0.75pt]
   RingMo \cite{sun2022ringmo} & 120 & 299 & 67.0\\
  \bottomrule[1.5pt]
  \end{tabular}}
\end{table}

\subsection{Remote Sensing Semantic Segmentation}

\begin{table*}[h!t]
  \caption{Comparisons with the SOTA Semantic Segmentation Methods on the Test Set of Potsdam.}
  \centering
  \label{tab6}
  \renewcommand\arraystretch{1.8}
  \setlength{\tabcolsep}{4.0mm}{
  \begin{tabular}{c|c|c|c|c|c|c|c}
  \toprule[1.5pt]
  
   \multirow{2}{*}{\textbf{Method}} & \multicolumn{6}{c|}{\textbf{IoU of each category(\%)}} &\multirow{2}{*}{\textbf{OA(\%)}} \\
   \cline{2-7}
   \multirow{2}{*}{} & \textbf{Imper.surf.} & \textbf{Building} & \textbf{Low veg.} & \textbf{Tree} & \textbf{Car} & \textbf{Clus.} & \multirow{2}{*}{}\\
   
   \midrule[0.75pt]
   
   PSPNet (r18) \cite{zhao2017pyramid} & 93.20 & 96.79 & 89.71 & 86.06 & 94.89 & 44.80 & 90.10 \\
   PSPNet (r50) \cite{zhao2017pyramid} & 93.68 & 97.41 & 89.81 & 86.48 & 96.02 & 46.73 & 90.60\\
   PSPNet (Swin Tiny) \cite{zhao2017pyramid} & 93.29 & 97.02 & 89.88 & \textbf{87.75} & 89.20 & 46.46 & 90.49 \\
   DeeplabV3+ (r18) \cite{chen2018encoder} & 93.09 & 97.26 & 89.82 & 85.65 & 95.81 & 44.36 & 90.13 \\
   DeeplabV3+ (r50) \cite{chen2018encoder} & 93.58 & 97.49 & 89.85 & 86.90 & 96.00 & \textbf{47.76} & 90.71\\
   DeeplabV3+ (Swin Tiny) \cite{chen2018encoder} & 94.12 & 97.49 & \textbf{91.30} & 85.69 & 96.22 & 45.50 & 90.88 \\
   \makecell[c]{RingMo-lite}  & \textbf{94.15} & \textbf{97.49} & 90.24 & 86.98 & \textbf{96.30} & 47.07 & \textbf{90.96} \\
   
  \bottomrule[1.5pt]
  \end{tabular}}
\end{table*}

\begin{table}[h!t]
  \caption{Comparisons of Params and FLOPs of Different Methods on Potsdam Dataset.}
  \centering
  \label{tab_potsdam_para}
  \renewcommand\arraystretch{2.3}
  \setlength{\tabcolsep}{2.7mm}{
  \begin{tabular}{c|c|c|c}
  \toprule[1.5pt]

   \textbf{Method} & \textbf{Para. (M)} & \textbf{FLOPs (G)} & \textbf{OA (\%)} \\
   \midrule[0.75pt]
   DeeplabV3+ (r50) \cite{chen2018encoder}  & 41.219 & 177 & 90.71\\
   \makecell[c]{DeeplabV3+\\ (Swin Tiny)(Supervised)}  & 41.885 & 38.419 & 90.88\\
   \makecell[c]{DeeplabV3+\\(Swin Tiny)(MIM)} & 41.885 & 38.419 & 90.92\\
   \makecell[c]{RingMo-lite} & 41.912 & 38.746 & \textbf{90.96}\\
   \midrule[0.75pt]
   RingMo \cite{sun2022ringmo} & 120 & 299 & 91.15\\
  \bottomrule[1.5pt]
  \end{tabular}}
\end{table}

\emph{1) Dataset Introduction:} We conduct experiments on the iSAID and ISPRS Potsdam datasets corresponding to the above two segmentation subtasks respectively, demonstrating the effectiveness and generalization ability of the proposed model.

\textbf{iSAID \cite{waqas2019isaid}.} The iSAID dataset stands as a semantic segmentation dataset derived from DOTA, which comprises 2806 aerial imagery samples spanning resolutions spanning from 800 × 800 to 4000 × 13000 pixels. These data are primarily sourced from Google Earth, supplemented by contributions from both the JL-1 and GF-2 satellites. In total, the dataset encompasses 655451 instances from 15 different object categories, while the remaining non-object pixels are designated as background. This dataset thus yields a cumulative of 16 categories. Within the ensemble of 2806 high-resolution images, the partitioning allocates 1411 images, while 458 images for training, 957 for validation, and the remaining images for testing. Notably, due to a lack of annotations for the test set, the testing in this study is conducted on the validation set.

\textbf{ISPRS Potsdam.} The dataset known as ISPRS Potsdam, made available by the ISPRS Commission WG II/4, offers 38 aerial images in high resolution with a pixel density of 0.5 meters. These images have a size of 6000 × 6000 pixels, with 24 images assigned as training subsets and the remaining 14 images as test subsets. Within this dataset, fine-grained annotations span 6 distinct categories, including impervious surfaces, building, low vegetation, tree, car, and clutter. For the purposes of experimentation, this section employs images composed of the Near-Infrared, Red and Green spectral bands as provided by the dataset.

\emph{2) Detailed Experimental Settings:} The GPU and Docker image are consistent to the prior experiments. Training parameters are tuned to encourage model convergence, with a learning rate of 0.0005, momentum at 0.99 and weight decay at 0.0001. We evaluate our models on the test part of the iSAID and Potsdam datasets, following the open-source project of MMSegmentation.

\begin{figure*}
\centering
	\includegraphics[scale=0.14]{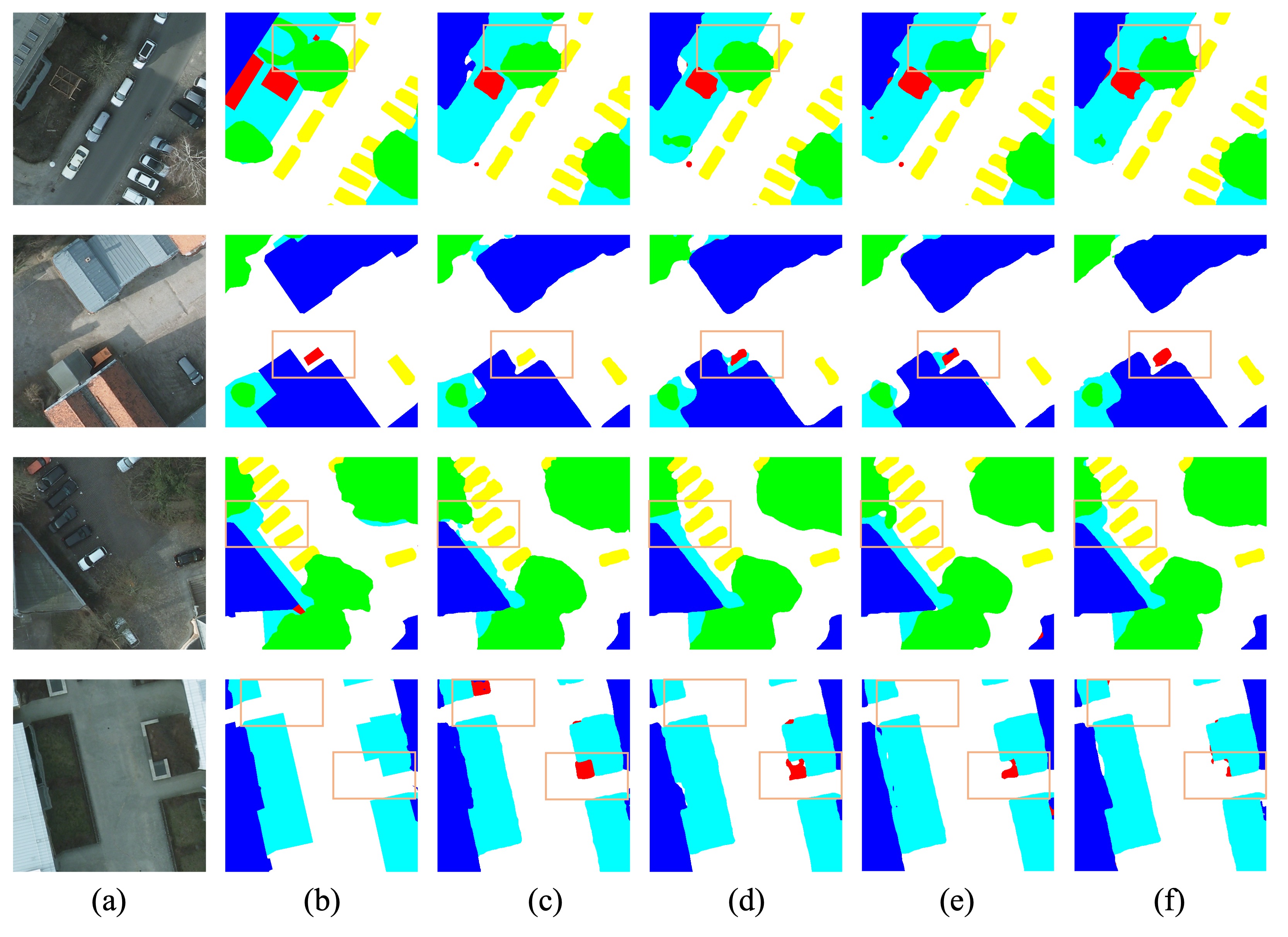}
\caption{Visualization results of the proposed RingMo-lite and the compared methods on the Potsdam dataset. (a) Images; (b) Labels; (c) PSPNet (r50); (d) DeeplabV3+ (r50); (e) DeeplabV3+ (Swin Tiny); (f) RingMo-lite. Our method achieves more accurate segmentation results compared to other methods.}
\label{fig_potsdam}
\end{figure*}

\emph{3) Experimental Result Analysis:} Table \ref{tab5} presents the segmentation accuracy results of our method and other state-of-the-art methods on the iSAID dataset. We use two different backbones of the ResNet series and the Swin Tiny series to complete the training of 80000 iterations on the iSAID dataset. Through the analysis of the results, it can be seen that compared with the ResNet series, Swin Tiny as a backbone has a better segmentation effect on the DeeplabV3+ segmentation method. In addition, we verify the effect of the FIFM module, and the results show that the method based on high and low frequencies has surpassed the effect of the current basic method (DeeplabV3+(r50)), and the improvement rate is 0.29\%.

Table \ref{tab6} shows the segmentation accuracy results of our method on the Potsdam dataset in comparison with other state-of-the-art methods. As in the Potsdam dataset, we use two different backbones to complete the training of the Potsdam dataset for 80000 iterations. Obviously, compared with using the ResNet series as the backbone, Swin Tiny as the backbone achieves a higher OA measure in the DeeplabV3+ method. Furthermore, the introduction of the FIFM module enables our method to outperform the current base method (DeeplabV3+ (r50)) by 0.25\%.

We compare the parameters and calculation amount of different methods, and the results of iSAID and Potsdam are presented in Table \ref{tab_isaid_para} and \ref{tab_potsdam_para}. It can be seen that the parameters of the RingMo-lite are only 34\% of the RingMo foundation model, and the OA is only reduced by 0.47\% and 0.19\%, respectively. Compared to the baselines of ResNet-50 and Swin Tiny, the parameters of RingMo-lite have increased slightly by 0.7M and 0.03M, respectively, while the effect has improved by 0.17\% and 0.12\% for iSAID, 0.21\% and 0.04\% for Potsdam. Finally, it is proved that RingMo-lite has an extreme low computation cost, which is only 12\% of the RingMo.

We show the visualization results on some samples in Potsdam in Fig. \ref{fig_potsdam}. It can be seen that RingMo-lite has a good segmentation effect on common RS scenes, from small cars to large constructions.

\subsection{Remote Sensing Change Detection}

Change detection stands as a critical task within the field of RS, holding significant implications for a multitude of applications. Its main objective revolves around discerning changes that occur within the same geographic range at different points in time. As a key technology for updating geographic data, Change detection plays an important role in assessing hazards and monitoring land cover.

\emph{1) Dataset Introduction:} In our pursuit to validate the efficacy of our proposed methodology, we embark upon experiments utilizing the LEVIR-CD \cite{2020A} dataset.

\textbf{LEVIR-CD} dataset is a novel large-scale binary change detection dataset in RS scenarios. Captured in Texas, USA, the dataset comprises a collection of bi-temporal images spanning timeframes from 2002 to 2018. The dataset consists of 637 pairs of images with very high resolutions (VHF), configured at dimensions of 1024 × 1024 pixels. The annotations provided in the LEVIR-CD dataset are in a binary format, specifically focusing on areas of change related to scenarios of building growth and building decline.

\emph{2) Detailed Experimental Settings:} The GPU and Docker image are consistent to the prior experiments. Training parameters are carefully adjusted to promote model convergence, including a learning rate of 0.0001, momentum set at 0.99, and weight decay of 0.01.  We evaluate our models on the test part of the LEVIR-CD dataset, following the set of RingMo.

\begin{table}[t]
  \caption{Comparisons of the SOTA Change Detection Methods on LEVIR-CD Dataset.}
  \centering
  \label{tab9}
  \renewcommand\arraystretch{1.8}
  \setlength{\tabcolsep}{4.0mm}{
  \begin{tabular}{c|c|c|c}
  \toprule[1.5pt]
  
   \multirow{2}{*}{\textbf{Method}} & \multicolumn{3}{c}{\textbf{LEVIR-CD}}  \\
   \cline{2-4}
   \multirow{2}{*}{} & \textbf{F1 (\%)} & \textbf{IoU (\%)} & \textbf{OA (\%)} \\
   
   \midrule[0.75pt]
   
    FC-EF \cite{daudt2018fully}   & 83.4 & 71.53 & 98.39 \\
    FC-Siam-Di \cite{daudt2018fully}   & 86.31 & 75.92 & 98.67 \\
    FC-Siam-Conc \cite{daudt2018fully}  & 83.69 & 71.96 & 98.49 \\
    DTCDSCN \cite{liu2020building}  & 87.67 & 78.05 & 98.77 \\
    STANet \cite{chen2020spatial} & 87.26 & 77.40 & 98.66 \\
    IFNet \cite{zhang2020deeply} & 88.13 & 78.77 & 98.87 \\
    SNUNet-CD \cite{fang2021snunet}  & 88.16 & 78.83 & 98.82 \\
    BiT \cite{wang2022empirical}  & 89.31 & 80.68 & 98.92 \\
    ChangeFormer \cite{bandara2022transformer}  & 90.40 & 82.48 & 99.04 \\
    DDPM-CD \cite{gedara2022remote}  & 90.91 & 83.35 & 99.09 \\
    RingMo-lite & \textbf{91.56} & \textbf{84.44} & \textbf{99.15} \\

  \bottomrule[1.5pt]
  \end{tabular}}

\end{table}

\begin{table}[h!t]
  \caption{Comparisons of Params and FLOPs of Different Methods on LEVIR-CD Dataset.}
  \centering
  \label{tab_levir_para}
  \renewcommand\arraystretch{2.3}
  \setlength{\tabcolsep}{2.7mm}{
  \begin{tabular}{c|c|c|c}
  \toprule[1.5pt]

   \textbf{Method} & \textbf{Para. (M)} & \textbf{FLOPs (G)} & \textbf{OA (\%)} \\
   \midrule[0.75pt]
   \makecell[c]{Di-Former \cite{chang2023transformer}\\ (Swin Tiny)(Supervised)}  & 54.26 & 16.84 & 99.12\\
   \makecell[c]{Di-Former \\(Swin Tiny)(MIM)} & 54.26 & 16.84 & 99.14\\
   \makecell[c]{RingMo-lite} & 54.38 & 16.84 & \textbf{99.15}\\
   \midrule[0.75pt]
   RingMo \cite{sun2022ringmo} & 122.1 & 95.5 & 99.18\\
  \bottomrule[1.5pt]
  \end{tabular}}
\end{table}

\emph{3) Experimental Result Analysis:} Table \ref{tab9} presents the change detection accuracy results of our method and other state-of-the-art methods on the LEVIR-CD dataset. We utilize CNN based methods like the FC series, DTCFSCN and the STANet, and the ViT based methods like the ChangeFormer, to complete the training of 200 epoches on the LEVIR-CD dataset. The analysis of the results reveals that Swin Tiny, as a backbone, outperforms both the ResNet series and ViT-based methods in terms of change detection effectiveness. In addition, we verify the effectiveness of the FIFM module, and the results show that the method based on high and low frequencies has surpassed the effect of the current baseline methods, resulting in an improvement rate of 0.06\% on OA.

We compare the parameters and calculation amount, and the results are presented in Table \ref{tab_levir_para}. Our method is based on DiFormer \cite{chang2023transformer}, which is based on Swin Transformer and has a differential feature enhancement module for RS change detection task. The parameters of the RingMo-lite are only 40\% of the RingMo foundation model, and the OA is only reduced by 0.03\%. Compared to the baselines of Swin Tiny, the parameters of RingMo-lite have increased slightly by 0.12M, while the effect has improved by 0.03\%. Finally, it is proved that RingMo-lite has an extreme low computation cost, which is only 17\% of the RingMo.

\subsection{Discussion}

Through a series of experiments and comparisons, we have demonstrated that RingMo-lite, with over 60\% fewer parameters than the original RingMo, exhibits only a marginal decrease (less than 2\%) in accuracy. Table \ref{tab_ringmolite_ringmo} emphasizes this advantages of RingMo-lite over the original RingMo. This achievement is attributed to the utilization of the FD-MIM pretraining method and the FIFB frequency-domain module, allowing RingMo-lite to approach the performance of large models more closely. Meanwhile, the performance across various tasks also demonstrates the generalization capability of RingMo-lite. It shows effectively extracting sample characteristics in multiple tasks. RingMo-lite will have significant relevance for lightweight foundation models in future RS tasks. However, RingMo-lite still faces several challenges.

\emph{\textbf{The performance of RingMo-lite is not impressive.}} RingMo-lite fails to surpass the performance of RingMo in all tasks. We plan to continue exploring the model structure to achieve higher accuracy.

\emph{\textbf{There is possible high load for devices.}} While RingMo-lite can be run on a single GPU during experiments, it still consumes a substantial amount of graphic memory. Its suitability for on-orbit scenario  requires further validation.

\emph{\textbf{RingMo-lite is not unified enough.}} RingMo-lite serves as a foundational model in the realm of pretraining, and it has not achieved unification in the field of lightweight models. Future research of lightweight multi-task RS models will be crucial.

\begin{table*}[h!t]
  \caption{Comparisons of RingMo-lite and RingMo in different tasks.}
  \centering
  \label{tab_ringmolite_ringmo}
  \renewcommand\arraystretch{1.8}
  \setlength{\tabcolsep}{3mm}{
  \begin{tabular}{c|c|c|c|c|c|c|c|c|c|c}
  \toprule[1.5pt]
  
   \multirow{2}{*}{\textbf{Tasks}} & \multirow{2}{*}{\textbf{Datasets}} & \multicolumn{3}{c|}{\textbf{RingMo-lite}} &
   \multicolumn{3}{c|}{\textbf{RingMo}} &
   \textbf{Para} &
   \textbf{FLOPs} &
   \textbf{Acc.} \\
   \cline{3-8}
   \multirow{2}{*}{} & \multirow{2}{*}{} & \textbf{Para.} & \textbf{FLOPs} & \textbf{Acc.} & \textbf{Para.} & \textbf{FLOPs} & \textbf{Acc.}  & \textbf{Ratio} & \textbf{Ratio} & \textbf{Diff.}\\
   
   \midrule[0.75pt]
   
   \multirow{5}{*}{\textbf{Classification}} & \textbf{UCM} & \multirow{5}{*}{28.294} & \multirow{5}{*}{4.494} & 99.05 & \multirow{5}{*}{87.768} & \multirow{5}{*}{15.438} & 99.06 &
    \multirow{5}{*}{32\%} & \multirow{5}{*}{29\%} & 0.01 \\
   \multirow{5}{*}{} & \textbf{AID-20\%} &  
   \multirow{5}{*}{} &  \multirow{5}{*}{} & 93.86 & 
   \multirow{5}{*}{} & \multirow{5}{*}{} & 96.27 &
    \multirow{5}{*}{} & \multirow{5}{*}{} & 2.41 \\
   \multirow{5}{*}{} & \textbf{AID-50\%} &  
   \multirow{5}{*}{} &  \multirow{5}{*}{} & 96.54 & 
   \multirow{5}{*}{} & \multirow{5}{*}{} & 98.13 &
    \multirow{5}{*}{} & \multirow{5}{*}{} & 1.59 \\
   \multirow{5}{*}{} & \textbf{NWPU-10\%} &  
   \multirow{5}{*}{} &  \multirow{5}{*}{} & 89.85 & 
   \multirow{5}{*}{} & \multirow{5}{*}{} & 93.01 &
    \multirow{5}{*}{} & \multirow{5}{*}{} & 3.16 \\
   \multirow{5}{*}{} & \textbf{AID-20\%} &  
   \multirow{5}{*}{} &  \multirow{5}{*}{} & 93.25 & 
   \multirow{5}{*}{} & \multirow{5}{*}{} & 95.28  & 
   \multirow{5}{*}{} & \multirow{5}{*}{} & 2.03\\
   
   \midrule[0.75pt]
   
   \textbf{Object} & \textbf{DIOR} & 37.36 & 196.07 & 73.4 & 97.14 & 450.66 & 74.7 &
   38\% & 43\% & 1.3 \\
   \textbf{Detection} & \textbf{FAIR1M} & 35.11 & 215.7 & 44.1 & 104.13 & 455.51 & 45.8 &
   33\% & 47\% & 1.7 \\
   
   \midrule[0.75pt]
   
   \textbf{Semantic} & \textbf{iSAID} & \multirow{2}{*}{41.912} & \multirow{2}{*}{38.746} & 66.53 & \multirow{2}{*}{120} & \multirow{2}{*}{299} & 67.0 & 
   \multirow{2}{*}{34\%} & \multirow{2}{*}{12\%} & 0.47 \\
   \textbf{Segmentation} & \textbf{Potsdam} & \multirow{2}{*}{} & \multirow{2}{*}{} & 90.96 & \multirow{2}{*}{} & \multirow{2}{*}{} & 91.15 &
   \multirow{2}{*}{} & \multirow{2}{*}{} & 0.19 \\
   
   \midrule[0.75pt]
   
   \textbf{Change Detection} & \textbf{LEVIR-CD} & 54.38 & 16.84 & 99.15 & 122.1 & 92.5 & 99.18 & 
   44\% & 17\% & 0.03 \\

   
  \bottomrule[1.5pt]
  \end{tabular}}
\end{table*}

\section{Conclusion}\label{Con}
This paper proposes RingMo-lite, a novel lightweight CNN-Transformer hybrid framework to address the challenge of efficient and accurate RS image interpretation on a lightweight platform. The framework considers the high-frequency and low-frequency domain characteristics of major target areas in different RS interpretation tasks and effectively balances global and local feature extraction. Self-supervised pretraining combined with pixel-level FD-MIM learns enriched feature representations from large amounts of data. The proposed method achieves SOTA performance in various RS tasks compared to models of the similar size. The proposed RingMo-lite reduces the parameters over 60\% compared to RingMo, and the average accuracy drops by less than 2\%. In addition, our work will be integrated into the MindSpore in the near future.

Moreover, we aim to incorporate richer patterns from various RS image modalities such as optical, SAR, infrared, and hyperspectral data. We will also expand our lightweight framework to cover a wider range of downstream applications, allowing the foundational model to capture more comprehensive and essential feature representations. Furthermore, we plan to design a lightweight foundation model suitable for multiple RS platforms and solve the challenges in RS image interpretation tasks more efficiently and accurately.

\bibliographystyle{IEEEtran}
\bibliography{refer}

\begin{thebibliography}{10}
\providecommand{\url}[1]{#1}
\csname url@samestyle\endcsname
\providecommand{\newblock}{\relax}
\providecommand{\bibinfo}[2]{#2}
\providecommand{\BIBentrySTDinterwordspacing}{\spaceskip=0pt\relax}
\providecommand{\BIBentryALTinterwordstretchfactor}{4}
\providecommand{\BIBentryALTinterwordspacing}{\spaceskip=\fontdimen2\font plus
\BIBentryALTinterwordstretchfactor\fontdimen3\font minus
  \fontdimen4\font\relax}
\providecommand{\BIBforeignlanguage}[2]{{%
\expandafter\ifx\csname l@#1\endcsname\relax
\typeout{** WARNING: IEEEtran.bst: No hyphenation pattern has been}%
\typeout{** loaded for the language `#1'. Using the pattern for}%
\typeout{** the default language instead.}%
\else
\language=\csname l@#1\endcsname
\fi
#2}}
\providecommand{\BIBdecl}{\relax}
\BIBdecl

\bibitem{liu2021swin}
Z.~Liu, Y.~Lin, Y.~Cao, H.~Hu, Y.~Wei, Z.~Zhang, S.~Lin, and B.~Guo, ``Swin
  transformer: Hierarchical vision transformer using shifted windows,'' in
  \emph{Proceedings of the IEEE/CVF international conference on computer
  vision}, 2021, pp. 10\,012--10\,022.

\bibitem{dong2023distilling}
Z.~Dong, G.~Gao, T.~Liu, Y.~Gu, and X.~Zhang, ``Distilling segmenters from cnns
  and transformers for remote sensing images semantic segmentation,''
  \emph{IEEE Transactions on Geoscience and Remote Sensing}, 2023.

\bibitem{chen2021remote}
H.~Chen, Z.~Qi, and Z.~Shi, ``Remote sensing image change detection with
  transformers,'' \emph{IEEE Transactions on Geoscience and Remote Sensing},
  vol.~60, pp. 1--14, 2021.

\bibitem{dosovitskiy2020image}
A.~Dosovitskiy, L.~Beyer, A.~Kolesnikov, D.~Weissenborn, X.~Zhai,
  T.~Unterthiner, M.~Dehghani, M.~Minderer, G.~Heigold, S.~Gelly \emph{et~al.},
  ``An image is worth 16x16 words: Transformers for image recognition at
  scale,'' \emph{arXiv preprint arXiv:2010.11929}, 2020.

\bibitem{deng2021joint}
W.~Deng, Q.~Liao, L.~Zhao, D.~Guo, G.~Kuang, D.~Hu, and L.~Liu, ``Joint
  clustering and discriminative feature alignment for unsupervised domain
  adaptation,'' \emph{IEEE Transactions on Image Processing}, vol.~30, pp.
  7842--7855, 2021.

\bibitem{sun2022ringmo}
X.~Sun, P.~Wang, W.~Lu, Z.~Zhu, X.~Lu, Q.~He, J.~Li, X.~Rong, Z.~Yang, H.~Chang
  \emph{et~al.}, ``Ringmo: A remote sensing foundation model with masked image
  modeling,'' \emph{IEEE Transactions on Geoscience and Remote Sensing}, 2022.

\bibitem{mehta2021mobilevit}
S.~Mehta and M.~Rastegari, ``Mobilevit: Light-weight, general-purpose, and
  mobile-friendly vision transformer,'' \emph{arXiv preprint arXiv:2110.02178},
  2021.

\bibitem{liu2021vision}
Y.~Liu, Y.-H. Wu, G.~Sun, L.~Zhang, A.~Chhatkuli, and L.~Van~Gool, ``Vision
  transformers with hierarchical attention,'' \emph{arXiv preprint
  arXiv:2106.03180}, 2021.

\bibitem{chen2022mobile}
Y.~Chen, X.~Dai, D.~Chen, M.~Liu, X.~Dong, L.~Yuan, and Z.~Liu,
  ``Mobile-former: Bridging mobilenet and transformer,'' in \emph{Proceedings
  of the IEEE/CVF Conference on Computer Vision and Pattern Recognition}, 2022,
  pp. 5270--5279.

\bibitem{cai2022efficientvit}
H.~Cai, C.~Gan, and S.~Han, ``Efficientvit: Enhanced linear attention for
  high-resolution low-computation visual recognition,'' \emph{arXiv preprint
  arXiv:2205.14756}, 2022.

\bibitem{pan2022edgevits}
J.~Pan, A.~Bulat, F.~Tan, X.~Zhu, L.~Dudziak, H.~Li, G.~Tzimiropoulos, and
  B.~Martinez, ``Edgevits: Competing light-weight cnns on mobile devices with
  vision transformers,'' in \emph{European Conference on Computer
  Vision}.\hskip 1em plus 0.5em minus 0.4em\relax Springer, 2022, pp. 294--311.

\bibitem{ma2022mocovit}
H.~Ma, X.~Xia, X.~Wang, X.~Xiao, J.~Li, and M.~Zheng, ``Mocovit: Mobile
  convolutional vision transformer,'' \emph{arXiv preprint arXiv:2205.12635},
  2022.

\bibitem{wu2022tinyvit}
K.~Wu, J.~Zhang, H.~Peng, M.~Liu, B.~Xiao, J.~Fu, and L.~Yuan, ``Tinyvit: Fast
  pretraining distillation for small vision transformers,'' in \emph{European
  Conference on Computer Vision}.\hskip 1em plus 0.5em minus 0.4em\relax
  Springer, 2022, pp. 68--85.

\bibitem{liu2022cross}
Y.~Liu, J.~Cao, B.~Li, W.~Hu, J.~Ding, and L.~Li, ``Cross-architecture
  knowledge distillation,'' in \emph{Proceedings of the Asian Conference on
  Computer Vision}, 2022, pp. 3396--3411.

\bibitem{chen2021glit}
B.~Chen, P.~Li, C.~Li, B.~Li, L.~Bai, C.~Lin, M.~Sun, J.~Yan, and W.~Ouyang,
  ``Glit: Neural architecture search for global and local image transformer,''
  in \emph{Proceedings of the IEEE/CVF International Conference on Computer
  Vision}, 2021, pp. 12--21.

\bibitem{wang2020hat}
H.~Wang, Z.~Wu, Z.~Liu, H.~Cai, L.~Zhu, C.~Gan, and S.~Han, ``Hat:
  Hardware-aware transformers for efficient natural language processing,''
  \emph{arXiv preprint arXiv:2005.14187}, 2020.

\bibitem{xu2021bert}
J.~Xu, X.~Tan, R.~Luo, K.~Song, J.~Li, T.~Qin, and T.-Y. Liu, ``Nas-bert:
  task-agnostic and adaptive-size bert compression with neural architecture
  search,'' in \emph{Proceedings of the 27th ACM SIGKDD Conference on Knowledge
  Discovery \& Data Mining}, 2021, pp. 1933--1943.

\bibitem{wadekar2022mobilevitv3}
S.~N. Wadekar and A.~Chaurasia, ``Mobilevitv3: Mobile-friendly vision
  transformer with simple and effective fusion of local, global and input
  features,'' \emph{arXiv preprint arXiv:2209.15159}, 2022.

\bibitem{zhang2022edgeformer}
H.~Zhang, W.~Hu, and X.~Wang, ``Edgeformer: Improving light-weight convnets by
  learning from vision transformers,'' \emph{arXiv preprint arXiv:2203.03952},
  vol.~2, 2022.

\bibitem{cheng2020remote}
G.~Cheng, X.~Xie, J.~Han, L.~Guo, and G.-S. Xia, ``Remote sensing image scene
  classification meets deep learning: Challenges, methods, benchmarks, and
  opportunities,'' \emph{IEEE Journal of Selected Topics in Applied Earth
  Observations and Remote Sensing}, vol.~13, pp. 3735--3756, 2020.

\bibitem{sun2020bas}
X.~Sun, A.~Shi, H.~Huang, and H.~Mayer, ``Bas$^{4}$net: Boundary-aware
  semi-supervised semantic segmentation network for very high resolution remote
  sensing images,'' \emph{IEEE Journal of Selected Topics in Applied Earth
  Observations and Remote Sensing}, vol.~13, pp. 5398--5413, 2020.

\bibitem{wang2023sar}
Z.~Wang, Y.~Kang, Z.~Xuan, Y.~Wang, T.~Zhang, and X.~Sun, ``Sar-aircraft-1.0:
  High-resolution sar aircraft detection and recognition dataset,''
  \emph{Journal of Radars}, vol.~12, no.~4, pp. 906--922, 2023.

\bibitem{xian2019air}
X.~Sun, Z.~Wang, Y.~Sun, W.~Diao, Y.~Zhang, and K.~Fu, ``Air-sarship-1.0:
  High-resolution sar ship detection dataset,'' \emph{Journal of Radars},
  vol.~8, no.~6, pp. 852--863, 2019.

\bibitem{lu2021lil}
X.~Lu, X.~Sun, W.~Diao, Y.~Feng, P.~Wang, and K.~Fu, ``Lil: Lightweight
  incremental learning approach through feature transfer for remote sensing
  image scene classification,'' \emph{IEEE Transactions on Geoscience and
  Remote Sensing}, vol.~60, pp. 1--20, 2021.

\bibitem{zhang2021cross}
Y.~Zhang, W.~Li, R.~Tao, J.~Peng, Q.~Du, and Z.~Cai, ``Cross-scene
  hyperspectral image classification with discriminative cooperative
  alignment,'' \emph{IEEE Transactions on Geoscience and Remote Sensing},
  vol.~59, no.~11, pp. 9646--9660, 2021.

\bibitem{zhao2020joint}
X.~Zhao, R.~Tao, W.~Li, H.-C. Li, Q.~Du, W.~Liao, and W.~Philips, ``Joint
  classification of hyperspectral and lidar data using hierarchical random walk
  and deep cnn architecture,'' \emph{IEEE Transactions on Geoscience and Remote
  Sensing}, vol.~58, no.~10, pp. 7355--7370, 2020.

\bibitem{sun2021pbnet}
X.~Sun, P.~Wang, C.~Wang, Y.~Liu, and K.~Fu, ``Pbnet: Part-based convolutional
  neural network for complex composite object detection in remote sensing
  imagery,'' \emph{ISPRS Journal of Photogrammetry and Remote Sensing}, vol.
  173, pp. 50--65, 2021.

\bibitem{zhu2020diamondnet}
Z.~Zhu, W.~Diao, K.~Chen, L.~Zhao, Z.~Yan, W.~Zhang, G.~Xu, and X.~Sun,
  ``Diamondnet: Ship detection in remote sensing images by extracting and
  clustering keypoints in a diamond,'' \emph{ISPRS Annals of the
  Photogrammetry, Remote Sensing and Spatial Information Sciences}, vol.~2, pp.
  625--632, 2020.

\bibitem{he2016deep}
K.~He, X.~Zhang, S.~Ren, and J.~Sun, ``Deep residual learning for image
  recognition,'' in \emph{Proceedings of the IEEE conference on computer vision
  and pattern recognition}, 2016, pp. 770--778.

\bibitem{sandler2018mobilenetv2}
M.~Sandler, A.~Howard, M.~Zhu, A.~Zhmoginov, and L.-C. Chen, ``Mobilenetv2:
  Inverted residuals and linear bottlenecks,'' in \emph{Proceedings of the IEEE
  conference on computer vision and pattern recognition}, 2018, pp. 4510--4520.

\bibitem{vaswani2017attention}
A.~Vaswani, N.~Shazeer, N.~Parmar, J.~Uszkoreit, L.~Jones, A.~N. Gomez,
  {\L}.~Kaiser, and I.~Polosukhin, ``Attention is all you need,''
  \emph{Advances in neural information processing systems}, vol.~30, 2017.

\bibitem{he2022masked}
K.~He, X.~Chen, S.~Xie, Y.~Li, P.~Doll{\'a}r, and R.~Girshick, ``Masked
  autoencoders are scalable vision learners,'' in \emph{Proceedings of the
  IEEE/CVF conference on computer vision and pattern recognition}, 2022, pp.
  16\,000--16\,009.

\bibitem{xie2022simmim}
Z.~Xie, Z.~Zhang, Y.~Cao, Y.~Lin, J.~Bao, Z.~Yao, Q.~Dai, and H.~Hu, ``Simmim:
  A simple framework for masked image modeling,'' in \emph{Proceedings of the
  IEEE/CVF Conference on Computer Vision and Pattern Recognition}, 2022, pp.
  9653--9663.

\bibitem{howard2017mobilenets}
A.~G. Howard, M.~Zhu, B.~Chen, D.~Kalenichenko, W.~Wang, T.~Weyand,
  M.~Andreetto, and H.~Adam, ``Mobilenets: Efficient convolutional neural
  networks for mobile vision applications,'' \emph{arXiv preprint
  arXiv:1704.04861}, 2017.

\bibitem{tan2019efficientnet}
M.~Tan and Q.~Le, ``Efficientnet: Rethinking model scaling for convolutional
  neural networks,'' in \emph{International conference on machine
  learning}.\hskip 1em plus 0.5em minus 0.4em\relax PMLR, 2019, pp. 6105--6114.

\bibitem{zhang2018shufflenet}
X.~Zhang, X.~Zhou, M.~Lin, and J.~Sun, ``Shufflenet: An extremely efficient
  convolutional neural network for mobile devices,'' in \emph{Proceedings of
  the IEEE conference on computer vision and pattern recognition}, 2018, pp.
  6848--6856.

\bibitem{cai2023efficientvit}
H.~Cai, J.~Li, M.~Hu, C.~Gan, and S.~Han, ``Efficientvit: Lightweight
  multi-scale attention for on-device semantic segmentation,'' 2023.

\bibitem{yang2022vitkd}
Z.~Yang, Z.~Li, A.~Zeng, Z.~Li, C.~Yuan, and Y.~Li, ``Vitkd: Practical
  guidelines for vit feature knowledge distillation,'' \emph{arXiv preprint
  arXiv:2209.02432}, 2022.

\bibitem{liang2023less}
C.~Liang, S.~Zuo, Q.~Zhang, P.~He, W.~Chen, and T.~Zhao, ``Less is more:
  Task-aware layer-wise distillation for language model compression,'' in
  \emph{International Conference on Machine Learning}.\hskip 1em plus 0.5em
  minus 0.4em\relax PMLR, 2023, pp. 20\,852--20\,867.

\bibitem{yu2022unified}
S.~Yu, T.~Chen, J.~Shen, H.~Yuan, J.~Tan, S.~Yang, J.~Liu, and Z.~Wang,
  ``Unified visual transformer compression,'' \emph{arXiv preprint
  arXiv:2203.08243}, 2022.

\bibitem{yin2023gohsp}
M.~Yin, B.~Uzkent, Y.~Shen, H.~Jin, and B.~Yuan, ``Gohsp: A unified framework
  of graph and optimization-based heterogeneous structured pruning for vision
  transformer,'' \emph{arXiv preprint arXiv:2301.05345}, 2023.

\bibitem{wei2023joint}
S.~Wei, T.~Ye, S.~Zhang, Y.~Tang, and J.~Liang, ``Joint token pruning and
  squeezing towards more aggressive compression of vision transformers,'' in
  \emph{Proceedings of the IEEE/CVF Conference on Computer Vision and Pattern
  Recognition}, 2023, pp. 2092--2101.

\bibitem{tan2019mnasnet}
M.~Tan, B.~Chen, R.~Pang, V.~Vasudevan, M.~Sandler, A.~Howard, and Q.~V. Le,
  ``Mnasnet: Platform-aware neural architecture search for mobile,'' in
  \emph{Proceedings of the IEEE/CVF conference on computer vision and pattern
  recognition}, 2019, pp. 2820--2828.

\bibitem{gong2021nasvit}
C.~Gong, D.~Wang, M.~Li, X.~Chen, Z.~Yan, Y.~Tian, V.~Chandra \emph{et~al.},
  ``Nasvit: Neural architecture search for efficient vision transformers with
  gradient conflict aware supernet training,'' in \emph{International
  Conference on Learning Representations}, 2021.

\bibitem{huang2023faster}
X.~Huang, F.~Liu, Y.~Cui, P.~Chen, L.~Li, and P.~Li, ``Faster and better: A
  lightweight transformer network for remote sensing scene classification,''
  \emph{Remote Sensing}, vol.~15, no.~14, p. 3645, 2023.

\bibitem{zheng2023lightweight}
F.~Zheng, S.~Lin, W.~Zhou, and H.~Huang, ``A lightweight dual-branch swin
  transformer for remote sensing scene classification,'' \emph{Remote Sensing},
  vol.~15, no.~11, p. 2865, 2023.

\bibitem{chen2023mdct}
J.~Chen, H.~Hong, B.~Song, J.~Guo, C.~Chen, and J.~Xu, ``Mdct: Multi-kernel
  dilated convolution and transformer for one-stage object detection of remote
  sensing images,'' \emph{Remote Sensing}, vol.~15, no.~2, p. 371, 2023.

\bibitem{gong2022swin}
H.~Gong, T.~Mu, Q.~Li, H.~Dai, C.~Li, Z.~He, W.~Wang, F.~Han, A.~Tuniyazi,
  H.~Li \emph{et~al.}, ``Swin-transformer-enabled yolov5 with attention
  mechanism for small object detection on satellite images,'' \emph{Remote
  Sensing}, vol.~14, no.~12, p. 2861, 2022.

\bibitem{wang2022unetformer}
L.~Wang, R.~Li, C.~Zhang, S.~Fang, C.~Duan, X.~Meng, and P.~M. Atkinson,
  ``Unetformer: A unet-like transformer for efficient semantic segmentation of
  remote sensing urban scene imagery,'' \emph{ISPRS Journal of Photogrammetry
  and Remote Sensing}, vol. 190, pp. 196--214, 2022.

\bibitem{xu2021efficient}
Z.~Xu, W.~Zhang, T.~Zhang, Z.~Yang, and J.~Li, ``Efficient transformer for
  remote sensing image segmentation,'' \emph{Remote Sensing}, vol.~13, no.~18,
  p. 3585, 2021.

\bibitem{park2022vision}
N.~Park and S.~Kim, ``How do vision transformers work?'' \emph{arXiv preprint
  arXiv:2202.06709}, 2022.

\bibitem{si2022inception}
C.~Si, W.~Yu, P.~Zhou, Y.~Zhou, X.~Wang, and S.~Yan, ``Inception transformer,''
  \emph{Advances in Neural Information Processing Systems}, vol.~35, pp.
  23\,495--23\,509, 2022.

\bibitem{xia2017aid}
G.-S. Xia, J.~Hu, F.~Hu, B.~Shi, X.~Bai, Y.~Zhong, L.~Zhang, and X.~Lu, ``Aid:
  A benchmark data set for performance evaluation of aerial scene
  classification,'' \emph{IEEE Transactions on Geoscience and Remote Sensing},
  vol.~55, no.~7, pp. 3965--3981, 2017.

\bibitem{cheng2017remote}
G.~Cheng, J.~Han, and X.~Lu, ``Remote sensing image scene classification:
  Benchmark and state of the art,'' \emph{Proceedings of the IEEE}, vol. 105,
  no.~10, pp. 1865--1883, 2017.

\bibitem{yang2010bag}
Y.~Yang and S.~Newsam, ``Bag-of-visual-words and spatial extensions for
  land-use classification,'' in \emph{Proceedings of the 18th SIGSPATIAL
  international conference on advances in geographic information systems},
  2010, pp. 270--279.

\bibitem{howard2019searching}
A.~Howard, M.~Sandler, G.~Chu, L.-C. Chen, B.~Chen, M.~Tan, W.~Wang, Y.~Zhu,
  R.~Pang, V.~Vasudevan, Q.~V. Le, and H.~Adam, ``Searching for mobilenetv3,''
  in \emph{2019 IEEE/CVF International Conference on Computer Vision (ICCV)},
  2019, pp. 1314--1324.

\bibitem{ma2018shufflenet}
N.~Ma, X.~Zhang, H.-T. Zheng, and J.~Sun, ``Shufflenet v2: Practical guidelines
  for efficient cnn architecture design,'' in \emph{Proceedings of the European
  conference on computer vision (ECCV)}, 2018, pp. 116--131.

\bibitem{ren2015faster}
S.~Ren, K.~He, R.~Girshick, and J.~Sun, ``Faster r-cnn: Towards real-time
  object detection with region proposal networks,'' \emph{Advances in neural
  information processing systems}, vol.~28, 2015.

\bibitem{lin2017focal}
T.-Y. Lin, P.~Goyal, R.~Girshick, K.~He, and P.~Doll{\'a}r, ``Focal loss for
  dense object detection,'' in \emph{Proceedings of the IEEE international
  conference on computer vision}, 2017, pp. 2980--2988.

\bibitem{yang2019reppoints}
Z.~Yang, S.~Liu, H.~Hu, L.~Wang, and S.~Lin, ``Reppoints: Point set
  representation for object detection,'' in \emph{Proceedings of the IEEE/CVF
  international conference on computer vision}, 2019, pp. 9657--9666.

\bibitem{cai2018cascade}
Z.~Cai and N.~Vasconcelos, ``Cascade r-cnn: Delving into high quality object
  detection,'' in \emph{Proceedings of the IEEE conference on computer vision
  and pattern recognition}, 2018, pp. 6154--6162.

\bibitem{ding2019learning}
J.~Ding, N.~Xue, Y.~Long, G.-S. Xia, and Q.~Lu, ``Learning roi transformer for
  oriented object detection in aerial images,'' in \emph{Proceedings of the
  IEEE/CVF Conference on Computer Vision and Pattern Recognition}, 2019, pp.
  2849--2858.

\bibitem{tian2019fcos}
Z.~Tian, C.~Shen, H.~Chen, and T.~He, ``Fcos: Fully convolutional one-stage
  object detection,'' in \emph{Proceedings of the IEEE/CVF international
  conference on computer vision}, 2019, pp. 9627--9636.

\bibitem{li2020object}
K.~Li, G.~Wan, G.~Cheng, L.~Meng, and J.~Han, ``Object detection in optical
  remote sensing images: A survey and a new benchmark,'' \emph{ISPRS journal of
  photogrammetry and remote sensing}, vol. 159, pp. 296--307, 2020.

\bibitem{sun2022fair1m}
X.~Sun, P.~Wang, Z.~Yan, F.~Xu, R.~Wang, W.~Diao, J.~Chen, J.~Li, Y.~Feng,
  T.~Xu \emph{et~al.}, ``Fair1m: A benchmark dataset for fine-grained object
  recognition in high-resolution remote sensing imagery,'' \emph{ISPRS Journal
  of Photogrammetry and Remote Sensing}, vol. 184, pp. 116--130, 2022.

\bibitem{kirillov2019panoptic}
A.~Kirillov, R.~Girshick, K.~He, and P.~Doll{\'a}r, ``Panoptic feature pyramid
  networks,'' in \emph{Proceedings of the IEEE/CVF conference on computer
  vision and pattern recognition}, 2019, pp. 6399--6408.

\bibitem{zhao2017pyramid}
H.~Zhao, J.~Shi, X.~Qi, X.~Wang, and J.~Jia, ``Pyramid scene parsing network,''
  in \emph{Proceedings of the IEEE conference on computer vision and pattern
  recognition}, 2017, pp. 2881--2890.

\bibitem{chen2018encoder}
L.-C. Chen, Y.~Zhu, G.~Papandreou, F.~Schroff, and H.~Adam, ``Encoder-decoder
  with atrous separable convolution for semantic image segmentation,'' in
  \emph{Proceedings of the European conference on computer vision (ECCV)},
  2018, pp. 801--818.

\bibitem{waqas2019isaid}
S.~Waqas~Zamir, A.~Arora, A.~Gupta, S.~Khan, G.~Sun, F.~Shahbaz~Khan, F.~Zhu,
  L.~Shao, G.-S. Xia, and X.~Bai, ``isaid: A large-scale dataset for instance
  segmentation in aerial images,'' in \emph{Proceedings of the IEEE/CVF
  Conference on Computer Vision and Pattern Recognition Workshops}, 2019, pp.
  28--37.

\bibitem{2020A}
H.~Chen and Z.~Shi, ``A spatial-temporal attention-based method and a new
  dataset for remote sensing image change detection,'' \emph{Remote Sensing},
  vol.~12, no.~10, p. 1662, 2020.

\bibitem{daudt2018fully}
R.~C. Daudt, B.~Le~Saux, and A.~Boulch, ``Fully convolutional siamese networks
  for change detection,'' in \emph{2018 25th IEEE International Conference on
  Image Processing (ICIP)}.\hskip 1em plus 0.5em minus 0.4em\relax IEEE, 2018,
  pp. 4063--4067.

\bibitem{liu2020building}
Y.~Liu, C.~Pang, Z.~Zhan, X.~Zhang, and X.~Yang, ``Building change detection
  for remote sensing images using a dual-task constrained deep siamese
  convolutional network model,'' \emph{IEEE Geoscience and Remote Sensing
  Letters}, vol.~18, no.~5, pp. 811--815, 2020.

\bibitem{chen2020spatial}
H.~Chen and Z.~Shi, ``A spatial-temporal attention-based method and a new
  dataset for remote sensing image change detection,'' \emph{Remote Sensing},
  vol.~12, no.~10, p. 1662, 2020.

\bibitem{zhang2020deeply}
C.~Zhang, P.~Yue, D.~Tapete, L.~Jiang, B.~Shangguan, L.~Huang, and G.~Liu, ``A
  deeply supervised image fusion network for change detection in high
  resolution bi-temporal remote sensing images,'' \emph{ISPRS Journal of
  Photogrammetry and Remote Sensing}, vol. 166, pp. 183--200, 2020.

\bibitem{fang2021snunet}
S.~Fang, K.~Li, J.~Shao, and Z.~Li, ``Snunet-cd: A densely connected siamese
  network for change detection of vhr images,'' \emph{IEEE Geoscience and
  Remote Sensing Letters}, vol.~19, pp. 1--5, 2021.

\bibitem{wang2022empirical}
D.~Wang, J.~Zhang, B.~Du, G.-S. Xia, and D.~Tao, ``An empirical study of remote
  sensing pretraining,'' \emph{IEEE Transactions on Geoscience and Remote
  Sensing}, 2022.

\bibitem{bandara2022transformer}
W.~G.~C. Bandara and V.~M. Patel, ``A transformer-based siamese network for
  change detection,'' in \emph{IGARSS 2022-2022 IEEE International Geoscience
  and Remote Sensing Symposium}.\hskip 1em plus 0.5em minus 0.4em\relax IEEE,
  2022, pp. 207--210.

\bibitem{gedara2022remote}
W.~Gedara Chaminda~Bandara, N.~Gopalakrishnan~Nair, and V.~M. Patel, ``Remote
  sensing change detection (segmentation) using denoising diffusion
  probabilistic models,'' \emph{arXiv e-prints}, pp. arXiv--2206, 2022.

\bibitem{chang2023transformer}
H.~Chang, X.~Sun, P.~Wang, W.~Diao, and G.~Xu, ``A transformer-based network
  with differential feature triple refinement for bitemporal remote sensing
  image change detection,'' 2023.

\end{thebibliography}

\vfill

\end{document}